\documentclass{article}

\usepackage{microtype}
\usepackage{graphicx}
\usepackage{subcaption}
\usepackage{booktabs} 

\usepackage{hyperref}

\usepackage{url}
\usepackage{multirow}
\usepackage{enumitem}
\usepackage{wrapfig}
\usepackage{booktabs}
\usepackage{subcaption}
\usepackage[most]{tcolorbox}
\usepackage{pgfplotstable}
\usepackage{adjustbox}
\usepackage{float}
\usepackage{hyperref}
\usepackage[perpage]{footmisc}
\usepackage{placeins}

\usepackage[preprint]{icml2026}

\usepackage{amsmath}
\usepackage{amssymb}
\usepackage{mathtools}
\usepackage{amsthm}

\usepackage[capitalize,noabbrev]{cleveref}

\theoremstyle{plain}

\theoremstyle{definition}

\theoremstyle{remark}

\usepackage[textsize=tiny]{todonotes}

\icmltitlerunning{Small Vectors, Big Effects: A Mechanistic Study of RL-Induced Reasoning via Steering Vectors}

\begin{document}

\twocolumn[
  \icmltitle{Small Vectors, Big Effects: \\ A Mechanistic Study of RL-Induced Reasoning via Steering Vectors}



  \icmlsetsymbol{equal}{*}

  \begin{icmlauthorlist}
    \icmlauthor{Viacheslav Sinii}{tink}
    \icmlauthor{Nikita Balagansky}{tink}
    \icmlauthor{Gleb Gerasimov}{tink}
    \icmlauthor{Daniil Laptev}{tink}
    \icmlauthor{Yaroslav Aksenov}{tink}
    \icmlauthor{Vadim Kurochkin}{tink}
    \icmlauthor{Alexey Gorbatovski}{tink}
    \icmlauthor{Boris Shaposhnikov}{tink}
    \icmlauthor{Daniil Gavrilov}{tink}
  \end{icmlauthorlist}

  \icmlaffiliation{tink}{T-Tech}

  \icmlcorrespondingauthor{Viacheslav Sinii}{v.siniy@t-tech.dev}

  \icmlkeywords{Machine Learning, ICML}

  \vskip 0.3in
]



\printAffiliationsAndNotice{}  


\begin{abstract}
The mechanisms by which reasoning training reshapes LLMs’ internal computations remain unclear. We study lightweight steering vectors inserted into the base model’s residual stream and trained with a reinforcement-learning objective. These vectors explain a large portion of full fine-tuning performance increase while preserving the interpretability of small, additive interventions. We find that (i) the last-layer steering vector acts like a token-substitution bias concentrated on the first generated token, consistently boosting tokens such as “To” and “Step”; (ii) the penultimate-layer vector leaves attention patterns largely intact and instead operates through the MLP and unembedding, preferentially up-weighting process words and structure symbols; and (iii) the steering vectors transfer to other models from the same family. Taken together, these results deepen understanding of how trained steering vectors shape computation and should inform future work in activation engineering and the study of reasoning models. The code is available at \url{https://github.com/corl-team/steering-reasoning}.
\end{abstract}
\section{Introduction}

Large reasoning models (LRMs) have recently shown remarkable performance \citep{jaech2024openai,guo2025deepseek} by being trained to produce effective chain-of-thoughts \citep{wei2022chain} before providing a final answer. Many open reproductions are trained on mathematical datasets with verifiable rewards \citep{hu2025open, liu2025oatzero}. However, we still lack a mechanistic understanding of the source of these gains.

Recent work in activation engineering \citep{turner2023steering, rimsky2024steering} has demonstrated that reasoning-connected behaviors such as backtracking are represented linearly inside the model and can be extracted from contrastive pairs \citep{venhoff2025understanding, ward2025reasoning}. Later, \citet{sinii2025steering} showed that trainable steering vectors can match the performance of fully fine-tuned models while involving a small, isolated set of parameters with the potential for interpretability. We build on this line of work by interpreting the effects of trainable steering vectors on LRMs’ behavior and on the circuitry they activate.

First, to isolate inter-layer effects, we train a single steering vector per model at a specific layer. Its performance provides an upper bound on what can be achieved by a linear intervention at that layer. We then present the following findings:
\begin{enumerate}[label=\textbullet, leftmargin=1.2em, labelsep=0.4em, itemsep=0pt, topsep=2pt, parsep=0pt]
\item \textbf{Early-layer steering does not express later-layer steering directions.} Although many layers achieve similar performance, the induced shifts diffuse as they propagate and become nearly orthogonal to later-layer steering vectors.
\item \textbf{Two mechanisms for steering.} The last-layer steering vector induces a shift in the output hidden states that is directionally dissimilar from the shifts induced by steering vectors at earlier layers.
\item \textbf{Last layer behaves like first-token substitution.} The final-layer vector acts at the unembedding, boosting opening tokens (e.g., ``To''/``Step''); simply prefixing these tokens recovers $\sim$10-11 points -- about three quarters of the explicit last-layer gain.
\item \textbf{Penultimate-layer vector acts through the MLP.} The induced effect is mediated almost entirely by the MLP, with minimal reliance on attention.
\item \textbf{Steering vectors transfer across related models.} Across multiple model families and sizes, steering vectors trained in one model often retain a non-trivial fraction of their performance when inserted into a closely matched model, suggesting that the underlying steering directions are largely preserved under fine-tuning and instruction tuning.

\end{enumerate}

\begin{figure*}[t] 
  \centering
  \begin{subfigure}{0.48\textwidth}
    \centering
    \includegraphics[width=\linewidth,page=1]{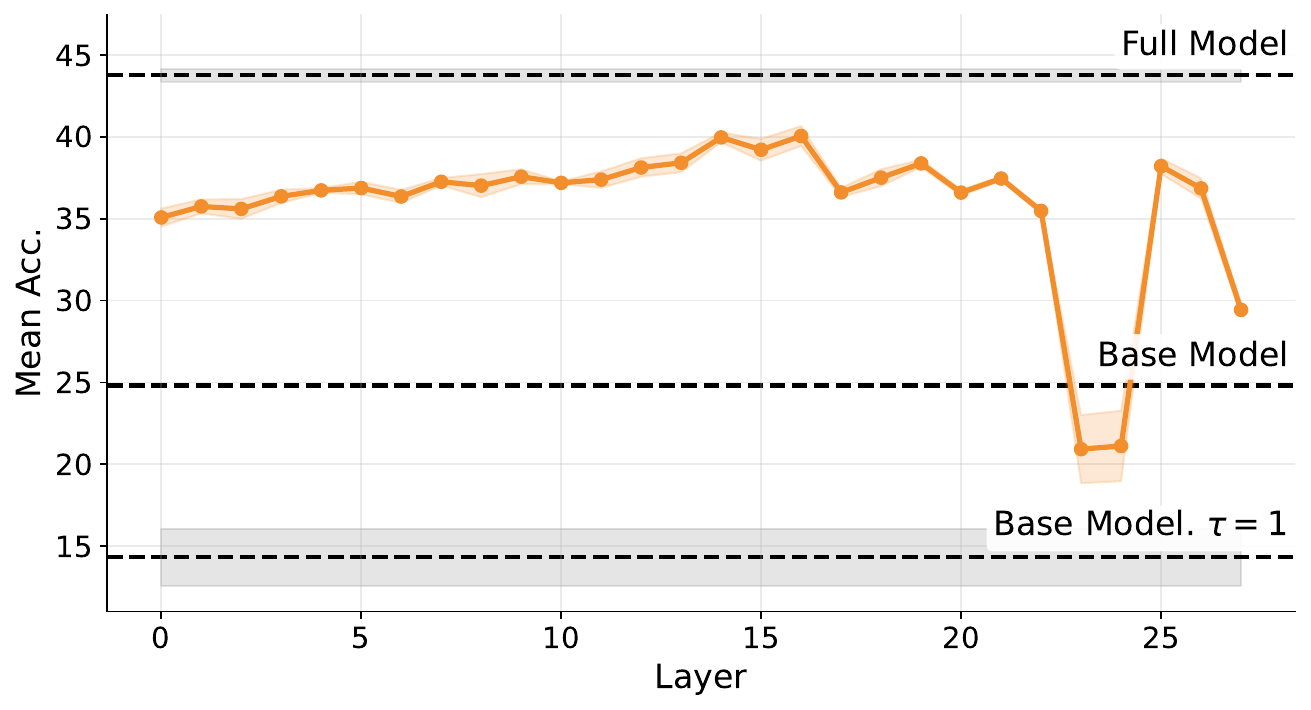}
    \caption*{Qwen2.5-Math-7B}
    \label{fig:per-layer-steering-qwen}
  \end{subfigure}\hfill
  \begin{subfigure}{0.48\textwidth}
    \centering
    \includegraphics[width=\linewidth,page=1]{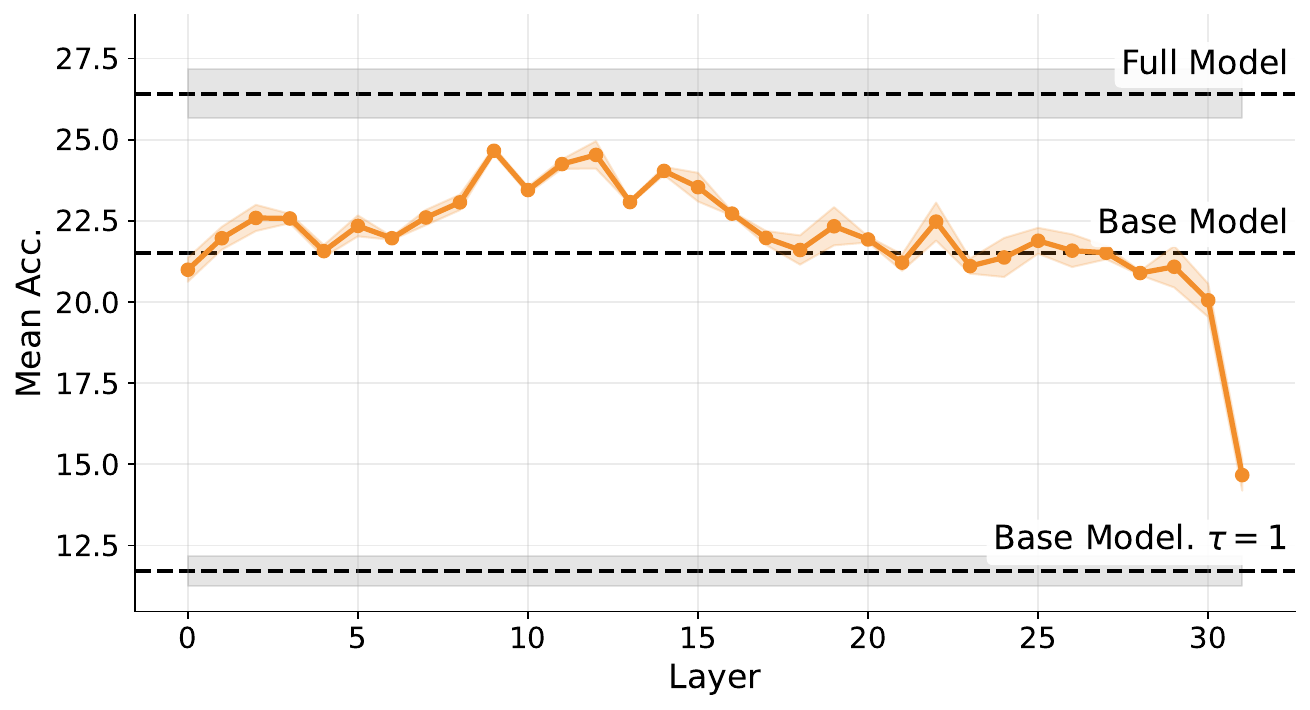}
    \caption*{LLaMa3.1-8B-It}
    \label{fig:per-layer-steering-llama}
  \end{subfigure}
  \caption{\textbf{Single-layer steering.} Mean accuracy on six benchmarks when training a single steering vector $s_\ell$ at layer $\ell$ (all other layers frozen). Many layers recover a substantial fraction of the gain from full fine-tuning, implying they captured reasoning-relevant information.}
  \label{fig:per-layer-steering}
\end{figure*}

\section{Background}
\label{sec:background}

Recent work has shown that training lightweight steering vectors can match the performance of fully tuned models \citep{sinii2025steering}. Concretely, zero-initialized steering vectors $s_\ell \in \mathbb{R}^d$ are added to the output residual stream of each layer $\ell$, while all other weights remain fixed. The vectors are trained with the RLOO \citep{ahmadian2024back} objective in a standard RLVR setup \citep{hu2025open, zeng2025simplerl, liu2025oatzero}. For a policy $\pi_\theta$, the policy-gradient update is
\[
\nabla_\theta J
= \mathbb{E}_{x \sim \mathcal{D},\, y \sim \pi_\theta(\cdot \mid x)}
\bigl[\, a(x,y)\, \nabla_\theta \log \pi_\theta(y \mid x) \,\bigr],
\]
where the advantage is defined as
\[
a(x,y) = r(x,y) - b(x),
\qquad
b(x) = \frac{1}{N}\sum_{y} r(x,y).
\]
Here $r(x,y)$ is the scalar reward for completion $y$ on prompt $x$, $N$ is the number of generated completions, and $b(x)$ is a per-prompt baseline used for variance reduction.

\citet{sinii2025steering} argue that this parameterization localizes training-induced changes in the model's internal computations, making the intervention easier to interpret. We adopt this setup to learn per-layer steering vectors for our interpretability study.

\begin{figure*}[t]
  \centering

  \begin{subfigure}[t]{\textwidth}
    \centering
    \begin{subfigure}[t]{0.48\textwidth}
      \centering
      \includegraphics[width=\linewidth,page=1]{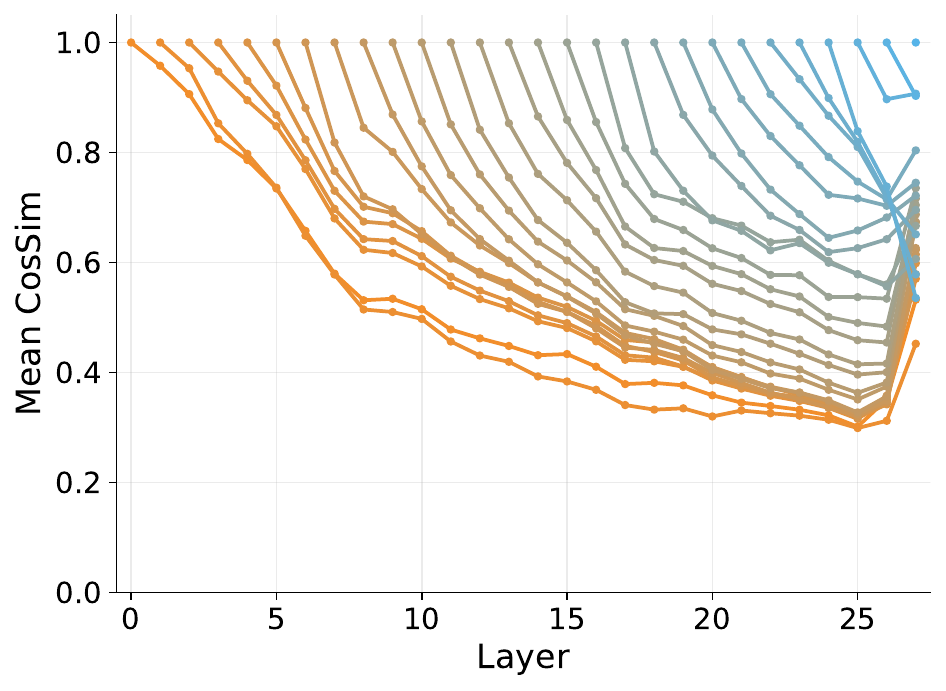}
      \caption*{Diff-Diff CosSim}
    \end{subfigure}\hfill
    \begin{subfigure}[t]{0.48\textwidth}
      \centering
      \includegraphics[width=\linewidth,page=1]{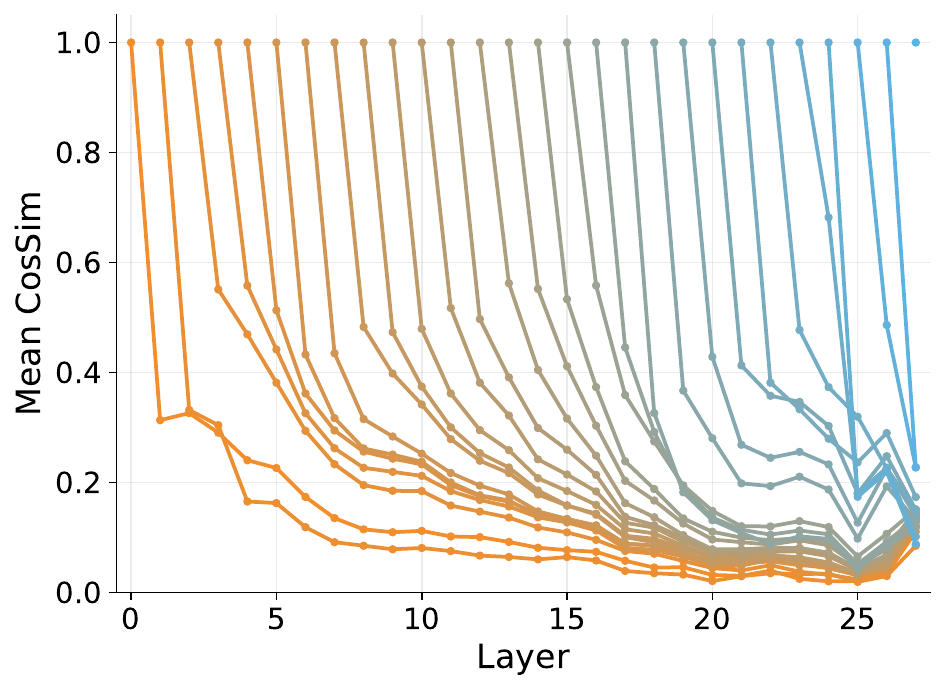}
      \caption*{Diff-Vector CosSim}
    \end{subfigure}
    \vspace{-1.0em}
    \caption*{\textbf{Qwen2.5-Math-7B}}
  \end{subfigure}

  \vspace{0.9em}

  \begin{subfigure}[t]{\textwidth}
    \centering

    \begin{subfigure}[t]{0.48\textwidth}
      \centering
      \includegraphics[width=\linewidth,page=1]{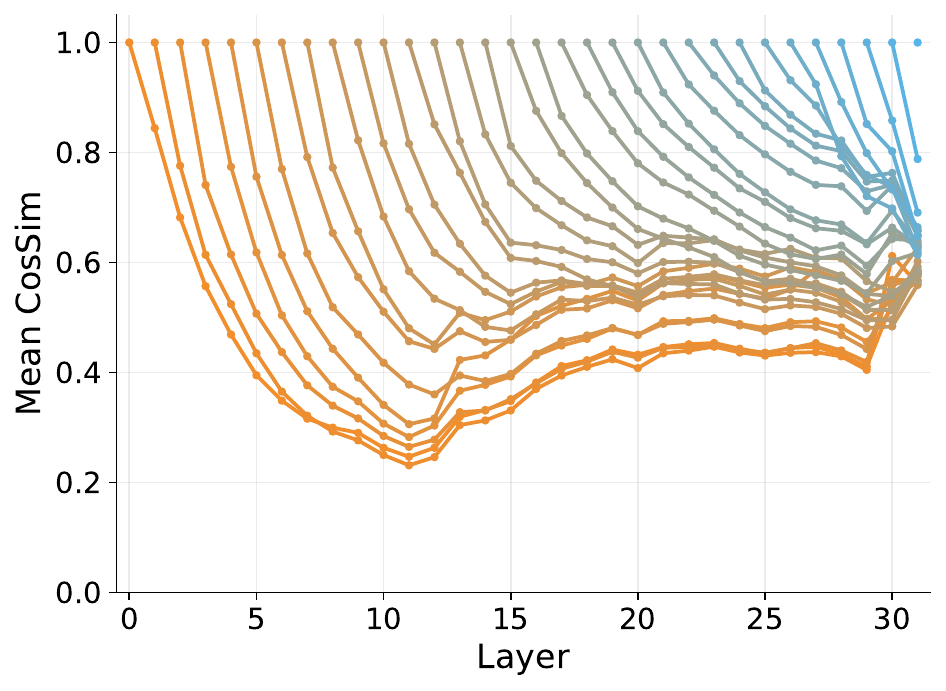}
      \caption*{Diff-Diff CosSim}
    \end{subfigure}\hfill
    \begin{subfigure}[t]{0.48\textwidth}
      \centering
      \includegraphics[width=\linewidth,page=1]{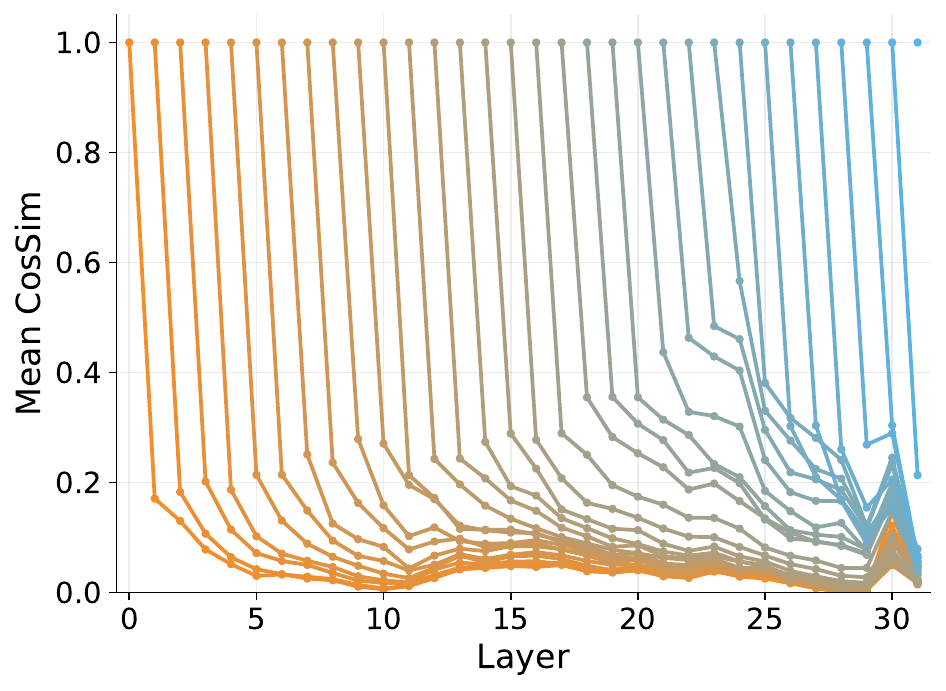}
      \caption*{Diff-Vector CosSim}
    \end{subfigure}
    \vspace{-1.0em}
    \caption*{\textbf{Llama3.1-8B-It}}
  \end{subfigure}

  \caption{
  \textbf{Steering Vector Persistence.}
For each steering layer \(i\) (color encodes \(i\); warm = early, cool = late) and each later layer \(\ell\) on the $x$-axis, we measure how similar the steering-induced hidden-state shift at layer $\ell$ is across tokens.
\emph{Left:} similarity between token's shift and the average shift over the dataset, quantifying how consistently the perturbation points in the same direction as it propagates.
\emph{Right:} similarity between each token's shift and the steering vector trained at layer $\ell$, quantifying whether the propagated effect aligns with that layer’s own steering direction.
}
  \label{fig:persistence_both}
\end{figure*}

\section{Single-layer steering vectors}
\label{sec:svs_match}

While all-layer steering vectors can match the performance of fully fine-tuned models, joint training couples layers through inter-layer interactions, making it difficult to attribute improvements to any single layer. To isolate layer-specific effects, we trained a steering vector at one layer at a time and report its performance. This per-layer setting both clarifies the object of our mechanistic analysis and provides a reference point -- an upper bound on what can be achieved by a single linear intervention at a fixed layer.

\paragraph{Setup.} In each training run, we selected a single layer $\ell$ and injected a steering vector $s_{\ell}$ into the residual stream at that layer’s output. All other layers remained unchanged. We initialized $s_{\ell}$ to zero and trained it using the standard RLVR pipeline.

We studied two base models -- Qwen2.5-Math-7B \citep{qwen25math} and Llama3.1-8B-It \citep{grattafiori2024llama}. Models were trained on the DeepScaleR dataset \citep{deepscaler2025} with the sampling temperature $\tau=1.0$, a 4K context window for Qwen2.5-Math-7B, and 8K for Llama3.1-8B-It. Rewards were assigned with \texttt{Math-Verify}\footnote{\url{https://github.com/huggingface/Math-Verify}}. We used $128$ prompts and $16$ generations per gradient step. We used the Adam optimizer \citep{kingma2014adam} with no weight decay. Evaluation spanned six math benchmarks: AIME24/25, AMC23, MATH500 \citep{hendrycks2021measuring}, MinervaMath \citep{lewkowycz2022solving}, and OlympiadBench \citep{he2024olympiadbench}. We report the mean score across these benchmarks in the main text and provide the raw numbers in \Cref{appendix:raw_bench_scores_layers}. For MATH500, MinervaMath, and OlympiadBench we report \textsc{Pass@1}; for AIME24/25 and AMC23 we report \textsc{Avg@32} due to their smaller sizes. During evaluation, models decoded with sampling at $\tau=1.0$ following \citet{zeng2025simplerl}. Evaluation context length was 4K and 32K for Qwen2.5-Math-7B and Llama3.1-8B-It, respectively. All metrics were averaged over three evaluation seeds. 

\paragraph{Result.} \Cref{fig:per-layer-steering} reports per-layer results for both models, compared with (i) all-layer steering, (ii) the base model with greedy decoding, and (iii) the base model sampled at $\tau=1.0$ (the training initialization). Most layers improve over the initialization, but none matches all-layer steering; under greedy decoding, several do (\Cref{appendix:per-layer-steering-temp-0}), suggesting that single-layer vectors target the right mechanisms yet cannot on their own sufficiently reduce the next-token distribution's entropy. In Qwen2.5-Math-7B, $s_{23}$ and $s_{24}$ underperform their neighboring layers; we trace the issue to vectors passing through the input layer-norm in layer $25$ (\Cref{appendix:bad_layers_qwen}).

\begin{figure}[t] 
  \centering
  \includegraphics[width=\columnwidth]{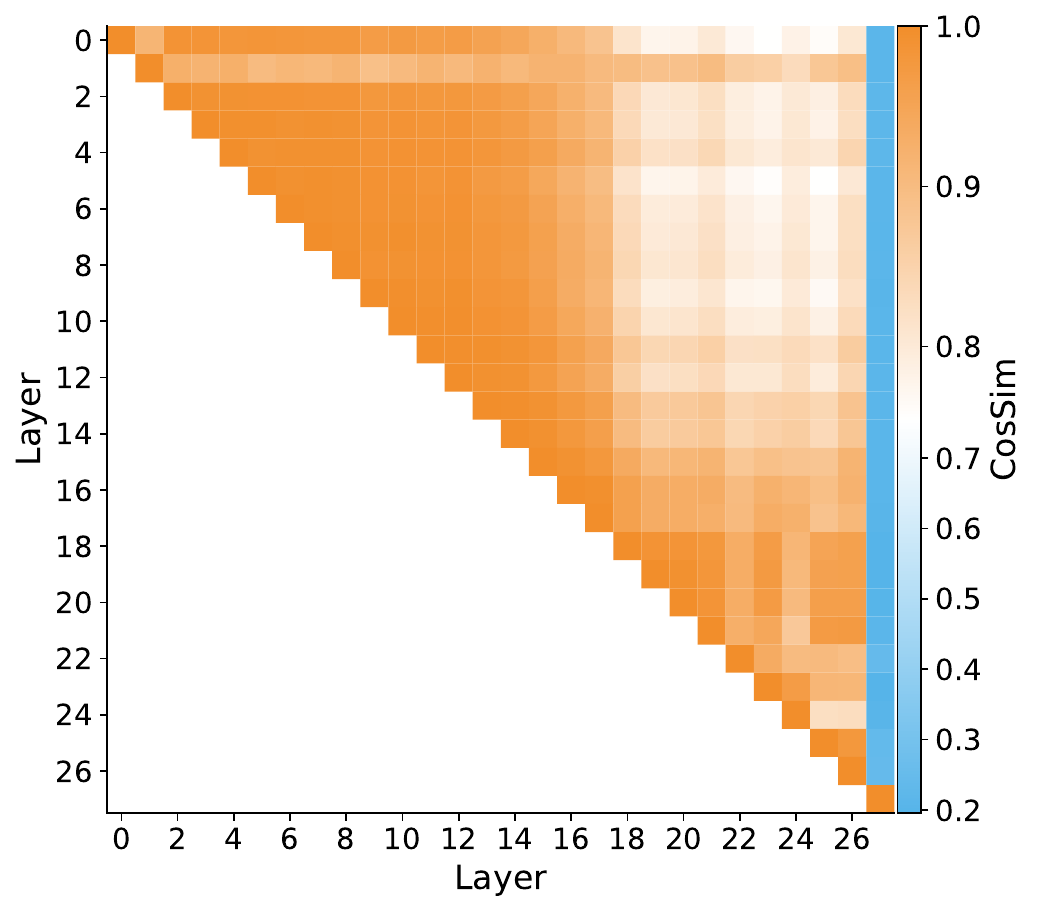}
\caption{
\textbf{Similarity of steering-induced unembedding biases (Qwen2.5-Math-7B).}
Each cell shows the cosine similarity between the average hidden-state shift at the final transformer layer induced by steering at layers $i$ and $j$. High similarity among $i,j<L$ indicates that steering from most layers produces a similar bias at the unembedding, largely independent of the injection point. In contrast, steering at the last layer directly yields a qualitatively different shift, implying a distinct mechanism.}
  \label{fig:persistence_last_diff_qwen}
\end{figure}%
\begin{figure*}[t]
  \centering

  \begin{subfigure}[t]{\textwidth}
    \centering
    \begin{subfigure}[t]{0.48\textwidth}
      \centering
      \includegraphics[width=\linewidth,page=1]{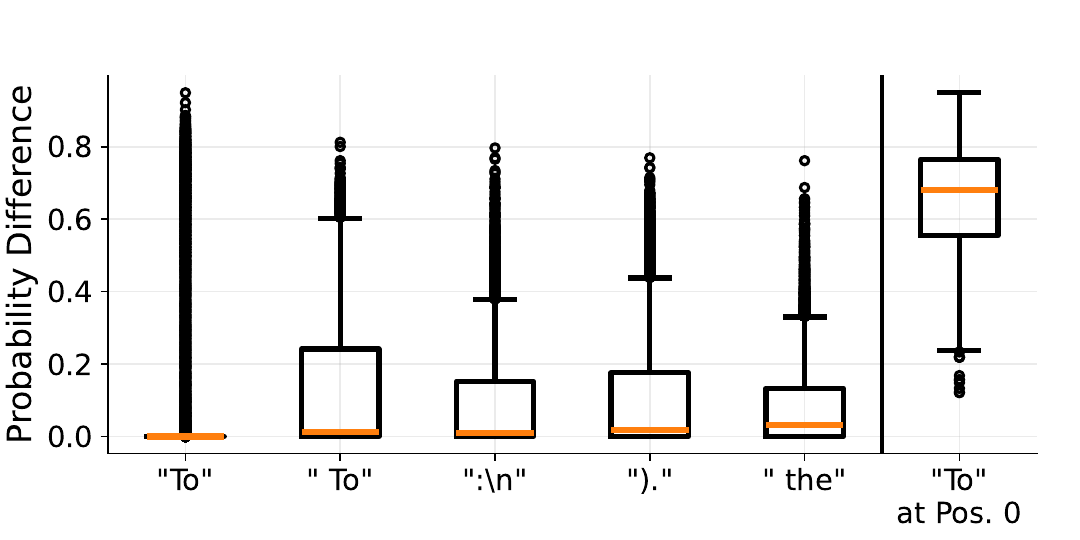}
      \caption*{}
    \end{subfigure}\hfill
    \begin{subfigure}[t]{0.48\textwidth}
      \centering
      \includegraphics[width=\linewidth,page=1]{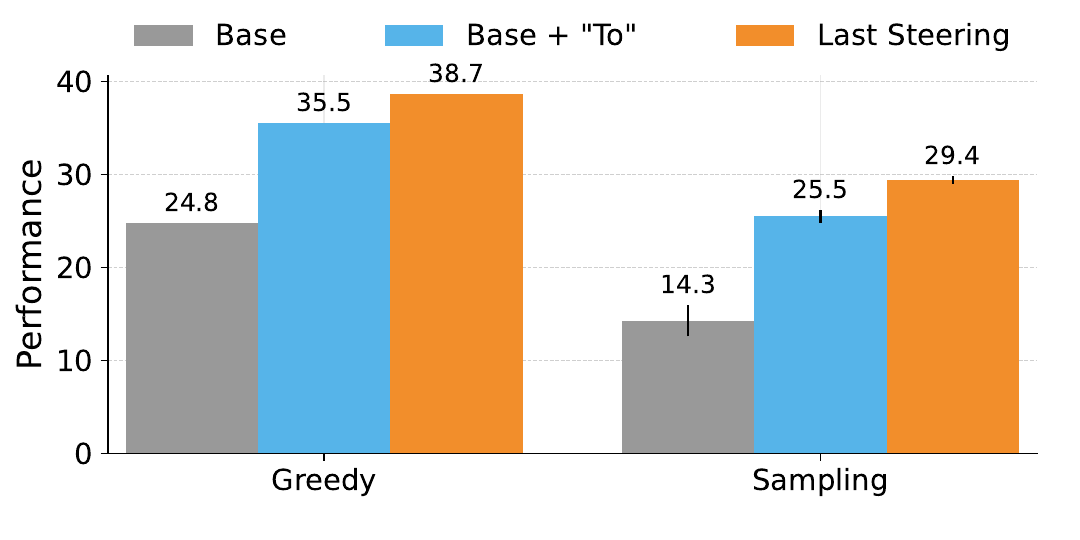}
      \caption*{}
    \end{subfigure}
    \vspace{-1.5em}
    \caption*{Qwen2.5-Math-7B}
  \end{subfigure}

  \vspace{0.9em}

  \begin{subfigure}[t]{\textwidth}
    \centering

    \begin{subfigure}[t]{0.48\textwidth}
      \centering
      \includegraphics[width=\linewidth,page=1]{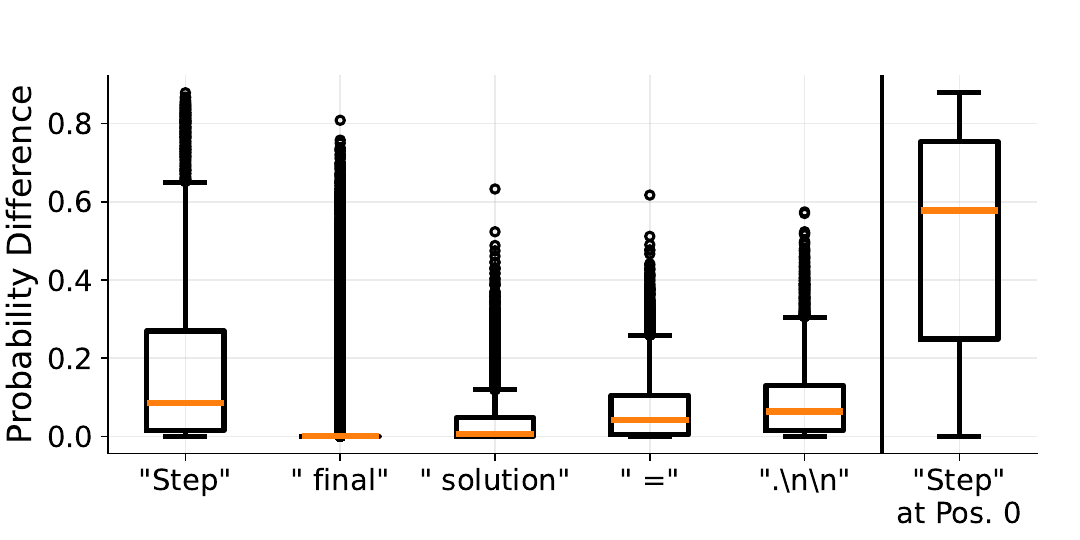}
      \caption*{}
    \end{subfigure}\hfill
    \begin{subfigure}[t]{0.48\textwidth}
      \centering
      \includegraphics[width=\linewidth,page=1]{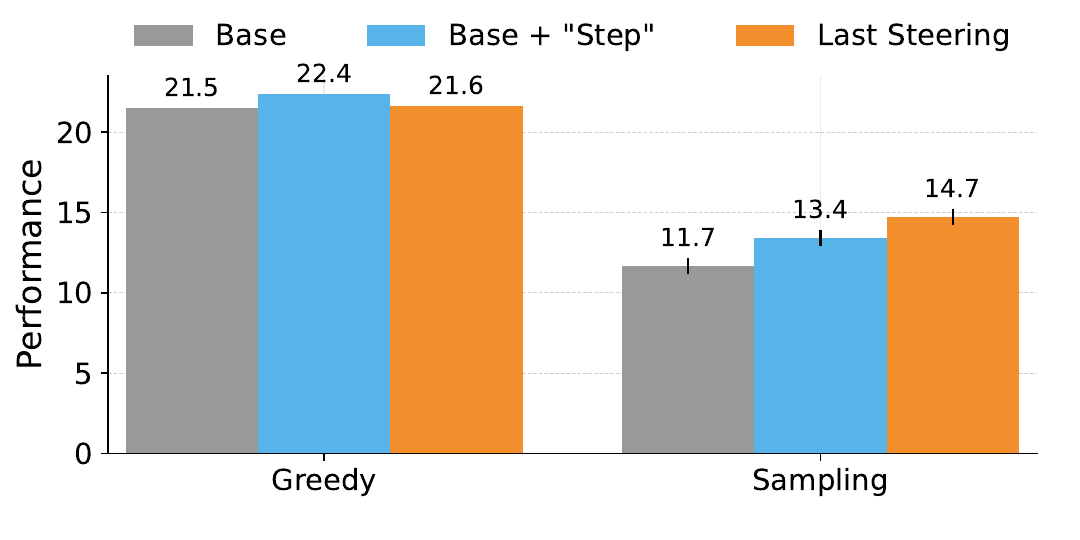}
      \caption*{}
    \end{subfigure}
    \vspace{-1.5em}
    \caption*{Llama3.1-8B-It}
  \end{subfigure}

  \caption{\textbf{Last-layer steering vector.}
\emph{Left}: distribution of token-level probability change induced by the last-layer vector over 1000 DeepScaleR prompts. We include the top-5 tokens by maximum change and highlight the most affected token at the zeroth generation position.
\emph{Right}: prefixing this token to each prompt reproduces a substantial fraction of the vector's accuracy gain under both greedy decoding and sampling.}
  \label{fig:token_to}
\end{figure*}

\section{Steering Vector Persistence}
\label{section:persistence}

\Cref{fig:per-layer-steering} shows that, in our experiments, steering at different layers yielded similar performance. One possible explanation is that an early steering vector could propagate through the network into a "virtual" linear steering vector at a later layer, targeting the same mechanism. We designed the next experiment to test this hypothesis.

For each input $x$, we computed the change in the layer-$\ell$ hidden state induced by steering at layer $i$:
\[
\Delta F_{<\ell,i}(x)
= F_{<\ell}(x;\,s_i) - F_{<\ell}(x),
\]
where $F_{<\ell}(x)$ is the output of the first $\ell$ layers of the transformer, and $s_i$ is the steering vector injected at layer $i$.

We then calculated two summary statistics:
\begin{enumerate}[leftmargin=1.2em, labelsep=0.4em, itemsep=0pt, topsep=2pt, parsep=0pt]
\item \textbf{Diff-Diff CosSim:} the average cosine similarity between $\Delta F_{<\ell,i}(x)$ and the mean effect $\mathbb{E}_x[\Delta F_{<\ell,i}(x)]$ over the dataset (how consistently the intervention pointed in the same direction).
\item \textbf{Diff-Vector CosSim:} the average cosine similarity between $\Delta F_{<\ell,i}(x)$ and the layer-$\ell$ steering vector $s_\ell$ (whether the propagated effect aligned with that layer's own steering direction).
\end{enumerate}

\Cref{fig:persistence_both} (top) shows the results for Qwen2.5-Math-7B.
\emph{Diff-Diff CosSim} indicates that (a) alignment of the induced shifts gradually decayed as the perturbation propagated; (b) the next layer received an almost uniform shift (cosine similarities were always $\ge 0.8$); (c) the shifts never became orthogonal (consistently $> 0.3$); and (d) alignment increased again in later layers. Taken together, these observations suggest that the induced shifts drifted as they traversed the network, but remained clustered around a common direction.

In contrast, \emph{Diff-Vector CosSim} dropped rapidly with distance from the injection layer: the propagated shifts were nearly orthogonal to each layer's own steering vector. Note that this alone does not imply different behaviors -- orthogonal steering directions could still induce the same behavior \citep{ifound800}. To test this possibility, we trained two steering vectors constrained to be orthogonal to the original vector at layers $14$ and $15$, and found that neither orthogonal vector matched the performance of the original one (\Cref{appendix:qwen_ortho}). Together with the \emph{Diff-Diff CosSim} results, these findings suggest that earlier-layer steering does not simply express the later-layer steering directions, and that steering at different layers relies on different learned mechanisms.

We observed a similar pattern for Llama3.1-8B-It (\Cref{fig:persistence_both}, bottom). The main difference was that \emph{Diff-Diff CosSim} reached its minimum in the middle layers and then rose toward later layers, suggesting a tighter concentration around a shared direction in the second half of the model. See \Cref{appendix:persistence_raw} for raw cosine similarity scores.

Finally, \emph{Diff-Diff CosSim} jumped at the final transformer layer (indexed as L), suggesting convergence to a more uniform effect at the unembedding. \Cref{fig:persistence_last_diff_qwen} shows pairwise cosine similarities between $\mathbb{E}_x[\Delta F_{<L,i}(x)]$ across injection layers, revealing roughly three direction groups. First, the last-layer steering effect was the most dissimilar from all others, with cosine similarity around $0.21$. Effects from other layers were much more similar, with a minimum cosine similarity of $0.75$, but still formed two clusters: layers $0$-$17$ and layers $18$-$26$. In the next two sections, we inspect two representative layers and study their behavioral effects on generation, as well as the mechanisms that produced them.

\begin{figure*}[t] 
  \centering
  \includegraphics[width=\linewidth,page=1]{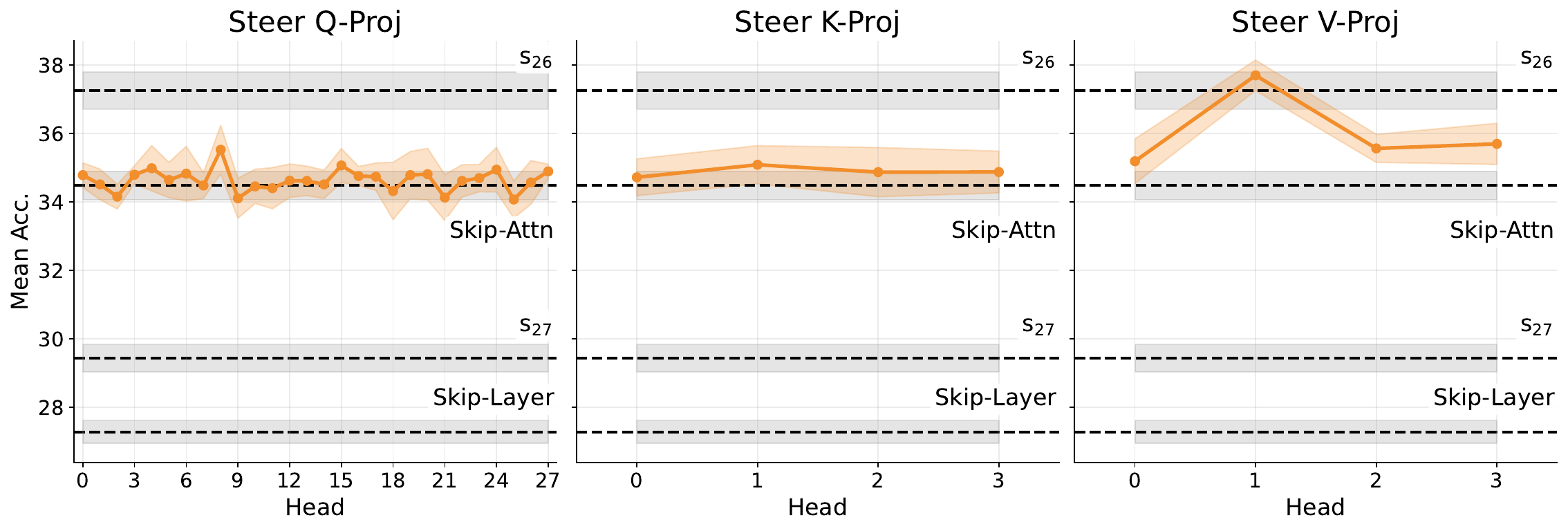}
  \caption{\textbf{Penultimate-layer steering in Qwen2.5-Math-7B.} Mean accuracy when injecting \(s_{26}\) into a single projection of the final block: ($Q$: left, $K$: center, $V$: right). Injecting only into value head $V_1$ closes the gap between \emph{Skip-Attn} and \(s_{26}\), indicating that the gain is carried by the value-output path and is largely independent of changes to the attention patterns.}
  \label{fig:patch_proj_qwen}
\end{figure*}

\begin{figure*}[t] 
  \centering
    \includegraphics[width=\linewidth]{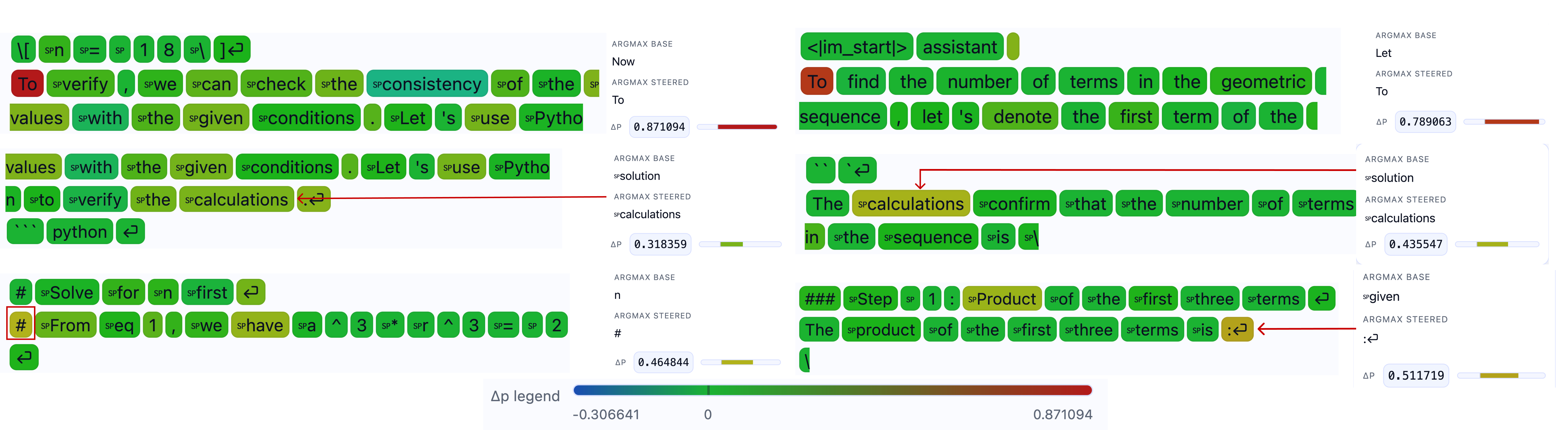}
  \caption{\textbf{Case study for Qwen2.5-Math-7B.} Token-level probability shifts (\(\Delta p\)) induced by \emph{penultimate-layer} steering. Three patterns emerge: \textbf{row 1} amplifies the first generated token \texttt{"To"}; \textbf{row 2} promotes process words (\texttt{"solution"} to \texttt{"calculations"}); \textbf{row 3} favors structural tokens that start Python code comments and newlines, instead of continuing the current sentence.}
  \label{fig:case_study}
\end{figure*}
\section{Last Layer -- Token Substitution}
\label{sec:last_layer}

\Cref{fig:per-layer-steering} shows that training only the last-layer vector $s_{27}$ in Qwen2.5-Math-7B closed over $50\%$ of the gap between the base model and full training. Since it accounts for a large part of the performance gain, we expected it to implement a simple and efficient strategy. With no subsequent layers to process it, $s_{27}$ acts directly at the unembedding and does not change earlier hidden states. This makes it behave like token substitution: it boosts logits of tokens it aligns with.

We read out these preferences using a \texttt{logit-lens} projection \citep{nostalgebraist2020lens}, multiplying $s_{27}$ by the unembedding matrix (omitting the pre-unembed layer norm). The top token was \emph{"To"}, with cosine similarity $0.37$; see \Cref{appendix:last_layer_logit_lens} for the top-10 tokens.

We next validated the behavioral impact on the model's generations. Although the vector is added unconditionally, the softmax nonlinearity means its effect depends on the initial logits. Let $x$ be the prefix (prompt plus the generated tokens so far). The next-token distribution is
\[
p(x) = \mathrm{Softmax}(g(F_{<L}(x))),
\]
where $g$ is the unembedding matrix, and $F_{<L}(x)$ is the final hidden state. With steering at the last layer, the induced change is
\[
\Delta p = p(x; s_L) - p(x).
\]
We estimated $\Delta p$ on 1000 DeepScaleR prompts, obtaining a score per token at each generation position.

Grouping by token, the largest increases were again for \texttt{"To"} and \texttt{"\ To"}, concentrated at the first generated token (\Cref{fig:token_to}, top-left). To test causality, we appended \texttt{"To"} to each prompt and evaluated the \emph{base} model: accuracy increased by 10--11 points under both greedy decoding and sampling, which is about $75\%$ of the gain from $s_{27}$ (\Cref{fig:token_to}, top-right).

The Llama3.1-8B-It results follow the same qualitative pattern as Qwen2.5-Math-7B, but the effect is weaker. In \Cref{fig:per-layer-steering}, the last-layer vector closes only a modest fraction of the gap to full training, and a \texttt{logit-lens} readout of $s_{31}$ shows weak alignment with any single token (max cosine similarity $0.12$; \Cref{appendix:last_layer_logit_lens}). Nonetheless, the highest-scoring tokens were variants of \texttt{"Step"} and \texttt{"final"}, and the induced probability changes were concentrated at the first generated position, primarily promoting \texttt{"Step"} (\Cref{fig:token_to}, bottom-left). Appending \texttt{"Step"} to each prompt improved the base model under both sampling and greedy decoding. Interestingly, under greedy decoding this prefix even outperformed last-layer steering, plausibly because a last-layer vector cannot condition its influence on position and thus also perturbs later steps.

Finally, we note that the same conclusions hold for models trained on a different dataset. In \Cref{appendix:token_to_opens1} we reran the experiment with models trained on open-s1 \citep{dang2025reinforcementlearningreasoningsmall}. Again, the last-layer vectors emphasized \texttt{"To"} (Qwen) and \texttt{"Step"} (Llama), primarily at the first generation position.

\begin{figure*}[t] 
  \centering
  \includegraphics[width=\linewidth,page=1]{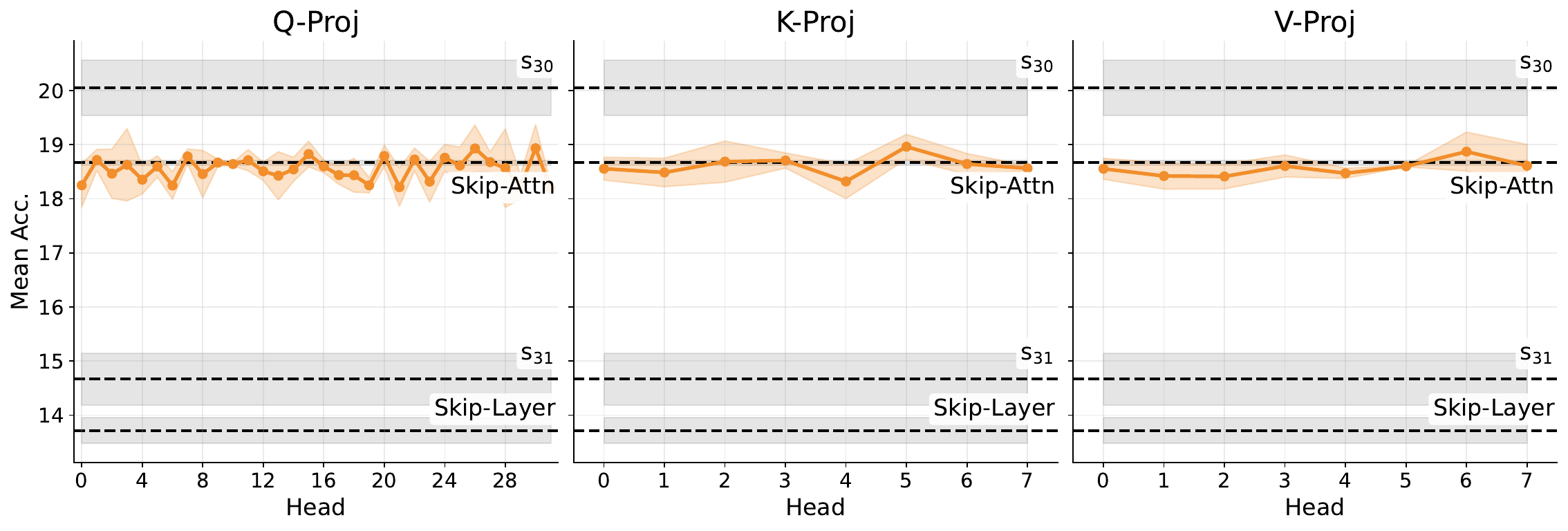}
  \caption{\textbf{Penultimate-layer steering in Llama3.1-8B-It.} Mean accuracy when injecting the penultimate-layer vector \(s_{30}\) into a single projection of the final block ($Q$: left, $K$: center, $V$: right). Injecting into any single projection remains close to \emph{Skip-Attn} and falls short of the full \(s_{30}\) result.}
  \label{fig:patch_proj_llama}
\end{figure*}

\begin{figure*}[t] 
  \centering
  \begin{subfigure}{0.48\textwidth}
    \centering
    \includegraphics[width=\linewidth]{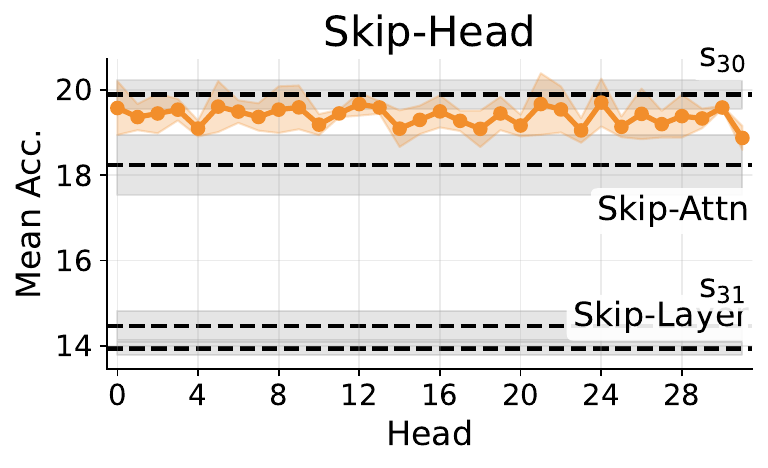}
  \end{subfigure}\hfill
  \begin{subfigure}{0.48\textwidth}
    \centering
    \includegraphics[width=\linewidth]{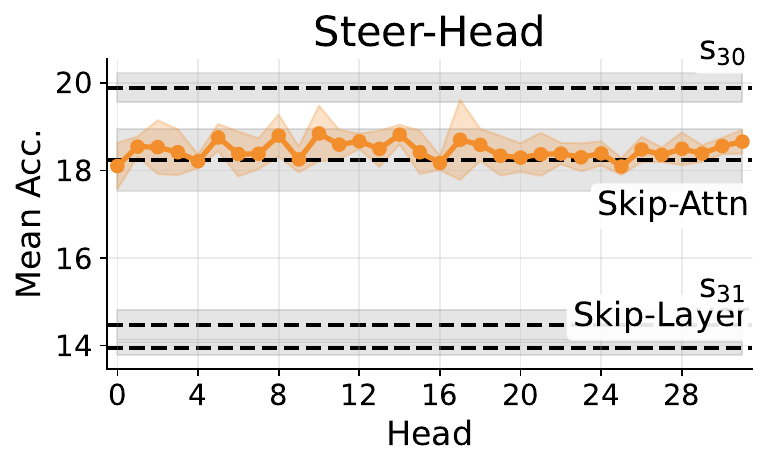}
  \end{subfigure}
  \caption{\textbf{Penultimate-layer steering in Llama3.1-8B-It.} Mean accuracy when patching whole heads in the final block with \(s_{30}\): \emph{Skip-Head} (left, steer all heads except head \(i\)) and \emph{Steer-Head} (right, steer only head \(i\)). No single head closes the gap between \emph{Skip-Attn} and \(s_{30}\), indicating a cooperative multi-head effect.}
  \label{fig:patch_head_llama}
\end{figure*}
\section{Penultimate Layer -- Circuit}
\label{section:pre_last_layer}

Steering the penultimate layer $s_{26}$ yielded a larger accuracy gain than steering the last layer, while remaining tractable to analyze because the modified activations traversed only one remaining block. Here we identify which parts of that block converted the steering signal into improved performance.

For a given residual input $X$, the layer computes $Y = X + \mathrm{MHA}(\mathrm{LN}(X))$ and $Z = Y + \mathrm{MLP}(\mathrm{LN}(Y))$, where $\mathrm{LN}$ denotes layer normalization. The Multi-Head Attention (MHA) mechanism consists of several attention heads, each defined as
\[
H_i(U) = \mathrm{Softmax}\!\left( \frac{U W_i^Q (U W_i^K)^\top}{\sqrt{d_k}} \right) U W_i^V,
\]
where $W_i^Q$, $W_i^K$, and $W_i^V$ are the query, key, and value projection matrices for head $i$. The outputs of all heads are concatenated and linearly transformed using an output projection matrix $W^O$, forming the complete attention output:
\[
\mathrm{MHA}(U) = [H_1(U); H_2(U); \ldots; H_h(U)] W^O.
\]
The Multi-Layer Perceptron (MLP) sublayer applies a two-layer feedforward transformation of the form
\[
\mathrm{MLP}(U) = f(U W_1 + b_1) W_2 + b_2,
\]
where $f(\cdot)$ is a nonlinear activation function. To understand the contribution of each submodule, we inserted or omitted the steering vector $s_{L-1}$ at specific points within the layer and measured the resulting change in mean accuracy. Specifically, we analyzed the following setups:
\begin{itemize}[leftmargin=1.2em, labelsep=0.4em, itemsep=0pt, topsep=2pt, parsep=0pt]
\item \textbf{Unmodified:} $X\!\leftarrow\!X+s_{L-1}$;
\item \textbf{Skip-Attn:} $Y\!\leftarrow\!Y+s_{L-1}$;
\item \textbf{Skip-Layer:} $Z\!\leftarrow\!Z+s_{L-1}$;
\item \textbf{Steer-Q/K/V-Proj:} for a head $i$, $(UW_i^{Q/K/V})\mapsto(U+s_{L-1})W_i^{Q/K/V}$.
\end{itemize}
If the change was significant, we marked the corresponding block as being important for processing the steering vector.

\Cref{fig:patch_proj_qwen} gives three takeaways for the Qwen2.5-Math-7B: (i) \emph{Skip-Layer} reduces accuracy relative to passing \(s_{26}\) through the block, showing an effect on the unembedding comparable to \(s_{27}\); (ii) \emph{Skip-Attn} preserves over half of the \(s_{26}\) gain, pointing to the MLP as the main contributor; (iii) patching any single \(Q\) or \(K\), or a \(V_j\) with \(j\neq1\), has little effect, whereas placing \(s_{26}\) only in \(V_1\) closes the gap to the full \(s_{26}\) result.

Since neither $Q$ nor $K$ projections are affected by the steering vector, the resulting QK circuit \citep{elhage2021mathematical} remains unchanged, preserving the attention pattern and the flow of information between tokens. Moreover, because \(s_{\,L-1}W^{V}_1W^{O}_1\) enters the residual regardless of attention weights (\Cref{appendix:v_1_proj}), this is equivalent to adding the projected vector just before the MLP, i.e., skipping attention. Indeed, a vector trained directly on the post-attention residual reached \(38.8\pm0.6\) mean accuracy, matching \(s_{26}\). Overall, the penultimate vector in Qwen2.5-Math-7B acts via two routes: a direct effect on the unembedding and an interaction with the MLP, largely bypassing attention.

\Cref{fig:case_study} shows how adding the steering vector to the post-attention residual stream shifts token probabilities. Beyond boosting the probability of the first token \texttt{"To"}, it promotes process words (e.g., replacing \texttt{"solution"} with \texttt{"calculations"}), possibly to deter premature endings. It also favors structural tokens such as Python comment markers and newlines, which often precede math blocks and may support in-code reasoning.

In contrast to Qwen, in Llama the projection-level patching (\textbf{Steer–Q/K/V}) did not reveal the source of the gain (\Cref{fig:patch_proj_llama}). We therefore went further and patched entire heads using two setups: 
\begin{itemize}[leftmargin=1.2em, labelsep=0.4em, itemsep=0pt, topsep=2pt, parsep=0pt]
\item \textbf{Steer-Head:} (\(H_i(U)\!\mapsto\! H_i(U+s_{\,L-1})\))
\item \textbf{Skip-Head:} (leave \(H_i(U)\) unchanged while steering all other heads).
\end{itemize}
In \Cref{fig:patch_head_llama}, two baselines mirror the Qwen result: \emph{Skip-Layer} performs close to \(s_{31}\), indicating a direct unembedding effect, and \emph{Skip-Attn} retains about \(70\%\) of the \(s_{30}\) gain, suggesting that much of the impact bypasses attention. No single head closes the remaining gap between \(s_{30}\) and \emph{Skip–Attn}, pointing to a cooperative multi-head mechanism and the importance of the attention layer for $s_{30}$'s performance. Still, training the steering vector directly in the post-attention residual stream yielded performance indistinguishable from \(s_{30}\) (mean accuracy \(19.9 \pm 0.1\)), suggesting either that the vector effectively bypasses attention or that attention contributes through a cooperative multi-head route.

\begin{table}[t]
\centering
\caption{\textbf{Transferability of steering vectors within model families.}
Each cell reports the normalized gain when steering vectors trained on the \emph{Donor} model are applied to the \emph{Recipient} model. 
Values are normalized such that the recipient equipped with its own vectors equals 1.0, and the base model (no vectors) equals 0.0; negative values indicate degradation. “---” denotes not applicable (no Math checkpoint available).}
\label{tab:transfer}

\begin{tabular}{llccc}
\toprule
\multicolumn{2}{c}{} & \multicolumn{3}{c}{\textbf{Donor}} \\
\cmidrule(lr){3-5}
\textbf{Family} & \textbf{Recipient} & \textbf{Base} & \textbf{Instruct} & \textbf{Math} \\
\midrule
\multirow{3}{*}{Qwen2.5-1.5B}
  & Base     & 1.00 & 0.38 & 0.32 \\
  & Instruct & 0.94 & 1.00 & 0.31 \\
  & Math     & 0.36 & 0.21 & 1.00 \\
\midrule
\multirow{3}{*}{Qwen2.5-7B}
  & Base     & 1.00 & 0.36 & 0.74 \\
  & Instruct & 0.55 & 1.00 & -0.34 \\
  & Math     & 0.32 & 0.05 & 1.00 \\
\midrule
\multirow{2}{*}{Llama-3.1-8B}
  & Base     & 1.00 & 0.28 & --- \\
  & Instruct & 0.74 & 1.00 & --- \\
\bottomrule
\end{tabular}
\end{table}

\section{Transfer of Steering Vectors Across Models}
\label{sec:transfer}

Finally, we tested whether the improvements induced by steering vectors transferred to other models within the same family. In this experiment, we used all-layer steering vectors, following the setup of \citet{sinii2025steering}. We considered three model groups: 
\begin{enumerate}[label=\textbullet, leftmargin=1.2em, labelsep=0.4em, itemsep=0pt, topsep=2pt, parsep=0pt]
\item \{Qwen2.5-7B, Qwen2.5-7B-Instruct, Qwen2.5-Math-7B\}
\item \{Qwen2.5-1.5B, Qwen2.5-1.5B-Instruct, Qwen2.5-Math-1.5B\}
\item \{Llama3.1-8B, Llama3.1-8B-Instruct\}
\end{enumerate}
where models within a group share hidden size and depth. For Llama3.1-8B-Instruct, we used the same chat template as the Llama3.1-8B.
For each ordered pair within a group, we swapped the donor model’s steering vectors into the recipient and report the \emph{relative} gain: scores are normalized by the gap between the base model and the same model equipped with its own vectors. Raw (unnormalized) scores are provided in \Cref{appendix:transfer_raw}.

\Cref{tab:transfer} summarizes the results. In most cases, the transfer yielded in a non-trivial gain, suggesting that the directions associated improved math performance are largely preserved after under fine-tuning and instruction tuning. Also, reusing steering vectors trained on a related base model can be a simple, low-cost way to improve performance on closely matched models.
\section{Related Work}
\label{appendix:related}

\textbf{Reinforcement learning with verifiable rewards.} \citet{jaech2024openai} demonstrated the striking performance of RL-tuned reasoning models, sparking a wave of follow-ups that develop these models \citep{guo2025deepseek, zeng2025simplerl, liu2025oatzero, hu2025open}. Subsequent work has examined why this training is effective, analysing model behaviour and the sources of its gains \citep{wang2025reinforcement, ye2025limo, shao2025spurious, liu2025understanding}. We contribute with a mechanistic study of the changes induced by reasoning training.

\textbf{Steering vectors} are small additive perturbations to the residual stream that modulate model behavior. They are widely viewed as \emph{feature amplifiers} -- strengthening existing computations rather than introducing new mechanisms -- and have been used to toggle or amplify \emph{reasoning-like} behaviors \citep{venhoff2025understanding, ward2025reasoning}. A common way to obtain them is \emph{contrastive extraction} from activation pairs (e.g., positive vs.\ negative sentiment) \citep{turner2023st_act_eng, panickssery2023st_contr, liu2023context, zou2023representation}. Beyond extraction, steering directions can also be \emph{trained}: optimized with preference data for controllable generation \citep{cao2024personalized}, or learned as simple additive vectors that surface latent behaviors such as step-by-step reasoning or self-reflection \citep{mack2024melbo, engels2025mechanisms, betley2025tell}.

In this work, we interpret steering vectors trained with GRPO-like objective using standard tools from mechanistic interpretability -- \texttt{logit-lens} to read out token-level effects \citep{nostalgebraist2020lens}, \emph{path patching} to localize circuits \citep{wang2022interpretability}, and circuit-style analyses in the QK/OV framework \citep{elhage2021mathematical}.

\section{Conclusion}
We presented a mechanistic interpretation of trainable steering vectors for mathematical reasoning. Across Qwen2.5-Math-7B and Llama3.1-8B-It, our results show three recurring patterns. First, last-layer steering behaves like first-token biasing at the unembedding: it preferentially promotes specific opening tokens (e.g., \texttt{"To"}/\texttt{"Step"}), and simple prefixing recovers a large fraction of the last-layer gain in Qwen and a smaller but consistent effect in Llama. Second, penultimate-layer steering achieves most of its benefit while largely bypassing attention: in Qwen the gain is carried primarily by a value-output path (with little dependence on attention-pattern changes), while in Llama no single projection or head explains the full effect, pointing to a cooperative multi-head contribution even though post-attention steering can match the penultimate-layer result. Third, steering effects do not propagate as a ``virtual'' later-layer direction: as perturbations travel forward, their directions diffuse and become nearly orthogonal to later-layer steering vectors, indicating distinct learned mechanisms across layers. In addition, our single-layer training setup provides a simple community baseline: training one vector at a time isolates layer-specific effects and enables reproducible comparisons to any linear intervention at the same layer across models and datasets.

Although most of these qualitative findings hold in both models, the effects are consistently stronger and cleaner in Qwen than in Llama (e.g., larger last-layer gains and clearer localization in the final block), suggesting model-specific implementations of similar high-level behaviors. Systematic comparisons across architectures and training recipes may help explain these differences. A promising direction is to pin down the precise mechanisms of mid-layer steering vectors, which often yield strong improvements but are harder to localize to a single submodule.

Overall, steering vectors provide a compact and informative probe of reasoning-trained models, offering concrete insight into how RL-induced changes manifest in token preferences and internal circuitry.

\section*{Impact Statement}
\label{sec:impact_statement}

This paper presents work whose goal is to advance the field of machine learning. There are many potential societal consequences of our work, none of which we feel must be specifically highlighted here

\bibliography{icml2026}

\begin{thebibliography}{36}
\providecommand{\natexlab}[1]{#1}
\providecommand{\url}[1]{\texttt{#1}}
\expandafter\ifx\csname urlstyle\endcsname\relax
  \providecommand{\doi}[1]{doi: #1}\else
  \providecommand{\doi}{doi: \begingroup \urlstyle{rm}\Url}\fi

\bibitem[Ahmadian et~al.(2024)Ahmadian, Cremer, Gall{\'e}, Fadaee, Kreutzer, Pietquin, {\"U}st{\"u}n, and Hooker]{ahmadian2024back}
Ahmadian, A., Cremer, C., Gall{\'e}, M., Fadaee, M., Kreutzer, J., Pietquin, O., {\"U}st{\"u}n, A., and Hooker, S.
\newblock Back to basics: Revisiting reinforce style optimization for learning from human feedback in llms.
\newblock \emph{arXiv preprint arXiv:2402.14740}, 2024.

\bibitem[Betley et~al.(2025)Betley, Bao, Soto, Sztyber-Betley, Chua, and Evans]{betley2025tell}
Betley, J., Bao, X., Soto, M., Sztyber-Betley, A., Chua, J., and Evans, O.
\newblock Tell me about yourself: Llms are aware of their learned behaviors.
\newblock \emph{arXiv preprint arXiv:2501.11120}, 2025.

\bibitem[Cao et~al.(2024)Cao, Zhang, Cao, Yin, Lin, Ma, and Chen]{cao2024personalized}
Cao, Y., Zhang, T., Cao, B., Yin, Z., Lin, L., Ma, F., and Chen, J.
\newblock Personalized steering of large language models: Versatile steering vectors through bi-directional preference optimization.
\newblock \emph{Advances in Neural Information Processing Systems}, 37:\penalty0 49519--49551, 2024.

\bibitem[Dang \& Ngo(2025)Dang and Ngo]{dang2025reinforcementlearningreasoningsmall}
Dang, Q.-A. and Ngo, C.
\newblock Reinforcement learning for reasoning in small llms: What works and what doesn't, 2025.
\newblock URL \url{https://arxiv.org/abs/2503.16219}.

\bibitem[Elhage et~al.(2021)Elhage, Nanda, Olsson, Henighan, Joseph, Mann, Askell, Bai, Chen, Conerly, et~al.]{elhage2021mathematical}
Elhage, N., Nanda, N., Olsson, C., Henighan, T., Joseph, N., Mann, B., Askell, A., Bai, Y., Chen, A., Conerly, T., et~al.
\newblock A mathematical framework for transformer circuits.
\newblock \emph{Transformer Circuits Thread}, 1\penalty0 (1):\penalty0 12, 2021.

\bibitem[Engels et~al.(2025)Engels, Nanda, and Rajamanoharan]{engels2025mechanisms}
Engels, J., Nanda, N., and Rajamanoharan, S.
\newblock Interim research report: Mechanisms of awareness.
\newblock \emph{AI Alignment Forum}, 2025.
\newblock \url{https://www.alignmentforum.org/posts/m8WKfNxp9eDLRkCk9/interim-research-report-mechanisms-of-awareness}.

\bibitem[Grattafiori et~al.(2024)Grattafiori, Dubey, Jauhri, Pandey, Kadian, Al-Dahle, Letman, Mathur, Schelten, Vaughan, et~al.]{grattafiori2024llama}
Grattafiori, A., Dubey, A., Jauhri, A., Pandey, A., Kadian, A., Al-Dahle, A., Letman, A., Mathur, A., Schelten, A., Vaughan, A., et~al.
\newblock The llama 3 herd of models.
\newblock \emph{arXiv preprint arXiv:2407.21783}, 2024.

\bibitem[Guo et~al.(2025)Guo, Yang, Zhang, Song, Zhang, Xu, Zhu, Ma, Wang, Bi, et~al.]{guo2025deepseek}
Guo, D., Yang, D., Zhang, H., Song, J., Zhang, R., Xu, R., Zhu, Q., Ma, S., Wang, P., Bi, X., et~al.
\newblock Deepseek-r1: Incentivizing reasoning capability in llms via reinforcement learning.
\newblock \emph{arXiv preprint arXiv:2501.12948}, 2025.

\bibitem[He et~al.(2024)He, Luo, Bai, Hu, Thai, Shen, Hu, Han, Huang, Zhang, et~al.]{he2024olympiadbench}
He, C., Luo, R., Bai, Y., Hu, S., Thai, Z.~L., Shen, J., Hu, J., Han, X., Huang, Y., Zhang, Y., et~al.
\newblock Olympiadbench: A challenging benchmark for promoting agi with olympiad-level bilingual multimodal scientific problems.
\newblock \emph{arXiv preprint arXiv:2402.14008}, 2024.

\bibitem[Hendrycks et~al.(2021)Hendrycks, Burns, Kadavath, Arora, Basart, Tang, Song, and Steinhardt]{hendrycks2021measuring}
Hendrycks, D., Burns, C., Kadavath, S., Arora, A., Basart, S., Tang, E., Song, D., and Steinhardt, J.
\newblock Measuring mathematical problem solving with the math dataset.
\newblock \emph{arXiv preprint arXiv:2103.03874}, 2021.

\bibitem[Hu et~al.(2025)Hu, Zhang, Han, Jiang, Zhang, and Shum]{hu2025open}
Hu, J., Zhang, Y., Han, Q., Jiang, D., Zhang, X., and Shum, H.-Y.
\newblock Open-reasoner-zero: An open source approach to scaling up reinforcement learning on the base model.
\newblock \emph{arXiv preprint arXiv:2503.24290}, 2025.

\bibitem[Jacob \& Turner(2024)Jacob and Turner]{ifound800}
Jacob, G.-W. and Turner, A.
\newblock I found \textgreater{}800 “orthogonal” write-code steering.
\newblock \url{https://www.lesswrong.com/posts/CbSEZSpjdpnvBcEvc/i-found-greater-than-800-orthogonal-write-code-steering}, 2024.
\newblock LessWrong. Accessed 2025-09-24.

\bibitem[Jaech et~al.(2024)Jaech, Kalai, Lerer, Richardson, El-Kishky, Low, Helyar, Madry, Beutel, Carney, et~al.]{jaech2024openai}
Jaech, A., Kalai, A., Lerer, A., Richardson, A., El-Kishky, A., Low, A., Helyar, A., Madry, A., Beutel, A., Carney, A., et~al.
\newblock Openai o1 system card.
\newblock \emph{arXiv preprint arXiv:2412.16720}, 2024.

\bibitem[Kingma(2014)]{kingma2014adam}
Kingma, D.~P.
\newblock Adam: A method for stochastic optimization.
\newblock \emph{arXiv preprint arXiv:1412.6980}, 2014.

\bibitem[Lewkowycz et~al.(2022)Lewkowycz, Andreassen, Dohan, Dyer, Michalewski, Ramasesh, Slone, Anil, Schlag, Gutman-Solo, et~al.]{lewkowycz2022solving}
Lewkowycz, A., Andreassen, A., Dohan, D., Dyer, E., Michalewski, H., Ramasesh, V., Slone, A., Anil, C., Schlag, I., Gutman-Solo, T., et~al.
\newblock Solving quantitative reasoning problems with language models.
\newblock \emph{Advances in neural information processing systems}, 35:\penalty0 3843--3857, 2022.

\bibitem[Liu et~al.(2023)Liu, Ye, Xing, and Zou]{liu2023context}
Liu, S., Ye, H., Xing, L., and Zou, J.
\newblock In-context vectors: Making in context learning more effective and controllable through latent space steering.
\newblock \emph{arXiv preprint arXiv:2311.06668}, 2023.

\bibitem[Liu et~al.(2025{\natexlab{a}})Liu, Chen, Li, Pang, Du, and Lin]{liu2025oatzero}
Liu, Z., Chen, C., Li, W., Pang, T., Du, C., and Lin, M.
\newblock There may not be aha moment in r1-zero-like training — a pilot study.
\newblock \url{https://oatllm.notion.site/oat-zero}, 2025{\natexlab{a}}.
\newblock Notion Blog.

\bibitem[Liu et~al.(2025{\natexlab{b}})Liu, Chen, Li, Qi, Pang, Du, Lee, and Lin]{liu2025understanding}
Liu, Z., Chen, C., Li, W., Qi, P., Pang, T., Du, C., Lee, W.~S., and Lin, M.
\newblock Understanding r1-zero-like training: A critical perspective.
\newblock \emph{arXiv preprint arXiv:2503.20783}, 2025{\natexlab{b}}.

\bibitem[Luo et~al.(2025)Luo, Tan, Wong, Shi, Tang, Roongta, Cai, Luo, Zhang, Li, Popa, and Stoica]{deepscaler2025}
Luo, M., Tan, S., Wong, J., Shi, X., Tang, W., Roongta, M., Cai, C., Luo, J., Zhang, T., Li, E., Popa, R.~A., and Stoica, I.
\newblock Deepscaler: Surpassing o1-preview with a 1.5b model by scaling rl.
\newblock \url{https://pretty-radio-b75.notion.site/DeepScaleR-Surpassing-O1-Preview-with-a-1-5B-Model-by-Scaling-RL-19681902c1468005bed8ca303013a4e2}, 2025.
\newblock Notion Blog.

\bibitem[Mack \& Turner(2024)Mack and Turner]{mack2024melbo}
Mack, A. and Turner, A.
\newblock Mechanistically eliciting latent behaviors in language models.
\newblock \emph{AI Alignment Forum}, 2024.
\newblock \url{https://www.alignmentforum.org/posts/ioPnHKFyy4Cw2Gr2x/mechanistically-eliciting-latent-behaviors-in-language-1}.

\bibitem[nostalgebraist(2020)]{nostalgebraist2020lens}
nostalgebraist.
\newblock interpreting gpt: the logit lens, 2020.
\newblock \url{https://www.alignmentforum.org/posts/AcKRB8wDpdaN6v6ru/interpreting-gpt-the-logit-lens}.

\bibitem[Panickssery et~al.(2023)Panickssery, Gabrieli, Schulz, Tong, Hubinger, and Turner]{panickssery2023st_contr}
Panickssery, N., Gabrieli, N., Schulz, J., Tong, M., Hubinger, E., and Turner, A.~M.
\newblock Steering llama 2 via contrastive activation addition.
\newblock \emph{arXiv preprint arXiv:2312.06681}, 2023.

\bibitem[Rimsky et~al.(2024)Rimsky, Gabrieli, Schulz, Tong, Hubinger, and Turner]{rimsky2024steering}
Rimsky, N., Gabrieli, N., Schulz, J., Tong, M., Hubinger, E., and Turner, A.
\newblock Steering llama 2 via contrastive activation addition.
\newblock In \emph{Proceedings of the 62nd Annual Meeting of the Association for Computational Linguistics (Volume 1: Long Papers)}, pp.\  15504--15522, 2024.

\bibitem[Shao et~al.(2025)Shao, Li, Xin, Geng, Wang, Oh, Du, Lambert, Min, Krishna, et~al.]{shao2025spurious}
Shao, R., Li, S.~S., Xin, R., Geng, S., Wang, Y., Oh, S., Du, S.~S., Lambert, N., Min, S., Krishna, R., et~al.
\newblock Spurious rewards: Rethinking training signals in rlvr.
\newblock \emph{arXiv preprint arXiv:2506.10947}, 2025.

\bibitem[Sinii et~al.(2025)Sinii, Gorbatovski, Cherepanov, Shaposhnikov, Balagansky, and Gavrilov]{sinii2025steering}
Sinii, V., Gorbatovski, A., Cherepanov, A., Shaposhnikov, B., Balagansky, N., and Gavrilov, D.
\newblock Steering llm reasoning through bias-only adaptation.
\newblock \emph{arXiv preprint arXiv:2505.18706}, 2025.

\bibitem[Turner et~al.(2023{\natexlab{a}})Turner, Thiergart, Leech, Udell, Vazquez, Mini, and MacDiarmid]{turner2023st_act_eng}
Turner, A.~M., Thiergart, L., Leech, G., Udell, D., Vazquez, J.~J., Mini, U., and MacDiarmid, M.
\newblock Steering language models with activation engineering.
\newblock \emph{arXiv preprint arXiv:2308.10248}, 2023{\natexlab{a}}.

\bibitem[Turner et~al.(2023{\natexlab{b}})Turner, Thiergart, Leech, Udell, Vazquez, Mini, and MacDiarmid]{turner2023steering}
Turner, A.~M., Thiergart, L., Leech, G., Udell, D., Vazquez, J.~J., Mini, U., and MacDiarmid, M.
\newblock Steering language models with activation engineering.
\newblock \emph{arXiv preprint arXiv:2308.10248}, 2023{\natexlab{b}}.

\bibitem[Venhoff et~al.(2025)Venhoff, Arcuschin, Torr, Conmy, and Nanda]{venhoff2025understanding}
Venhoff, C., Arcuschin, I., Torr, P., Conmy, A., and Nanda, N.
\newblock Understanding reasoning in thinking language models via steering vectors.
\newblock \emph{arXiv preprint arXiv:2506.18167}, 2025.

\bibitem[Wang et~al.(2022)Wang, Variengien, Conmy, Shlegeris, and Steinhardt]{wang2022interpretability}
Wang, K., Variengien, A., Conmy, A., Shlegeris, B., and Steinhardt, J.
\newblock Interpretability in the wild: a circuit for indirect object identification in gpt-2 small.
\newblock \emph{arXiv preprint arXiv:2211.00593}, 2022.

\bibitem[Wang et~al.(2025)Wang, Yang, Zeng, Ren, Liu, Peng, Cheng, He, Wang, Gao, et~al.]{wang2025reinforcement}
Wang, Y., Yang, Q., Zeng, Z., Ren, L., Liu, L., Peng, B., Cheng, H., He, X., Wang, K., Gao, J., et~al.
\newblock Reinforcement learning for reasoning in large language models with one training example.
\newblock \emph{arXiv preprint arXiv:2504.20571}, 2025.

\bibitem[Ward et~al.(2025)Ward, Lin, Venhoff, and Nanda]{ward2025reasoning}
Ward, J., Lin, C., Venhoff, C., and Nanda, N.
\newblock Reasoning-finetuning repurposes latent representations in base models.
\newblock \emph{arXiv preprint arXiv:2507.12638}, 2025.

\bibitem[Wei et~al.(2022)Wei, Wang, Schuurmans, Bosma, Xia, Chi, Le, Zhou, et~al.]{wei2022chain}
Wei, J., Wang, X., Schuurmans, D., Bosma, M., Xia, F., Chi, E., Le, Q.~V., Zhou, D., et~al.
\newblock Chain-of-thought prompting elicits reasoning in large language models.
\newblock \emph{Advances in neural information processing systems}, 35:\penalty0 24824--24837, 2022.

\bibitem[Yang et~al.(2024)Yang, Zhang, Hui, Gao, Yu, Li, Liu, Tu, Zhou, Lin, Lu, Xue, Lin, Liu, Ren, and Zhang]{qwen25math}
Yang, A., Zhang, B., Hui, B., Gao, B., Yu, B., Li, C., Liu, D., Tu, J., Zhou, J., Lin, J., Lu, K., Xue, M., Lin, R., Liu, T., Ren, X., and Zhang, Z.
\newblock Qwen2.5-math technical report: Toward mathematical expert model via self-improvement.
\newblock \emph{arXiv preprint arXiv:2409.12122}, 2024.

\bibitem[Ye et~al.(2025)Ye, Huang, Xiao, Chern, Xia, and Liu]{ye2025limo}
Ye, Y., Huang, Z., Xiao, Y., Chern, E., Xia, S., and Liu, P.
\newblock Limo: Less is more for reasoning.
\newblock \emph{arXiv preprint arXiv:2502.03387}, 2025.

\bibitem[Zeng et~al.(2025)Zeng, Huang, Liu, Liu, He, Ma, and He]{zeng2025simplerl}
Zeng, W., Huang, Y., Liu, Q., Liu, W., He, K., Ma, Z., and He, J.
\newblock Simplerl-zoo: Investigating and taming zero reinforcement learning for open base models in the wild.
\newblock \emph{arXiv preprint arXiv:2503.18892}, 2025.

\bibitem[Zou et~al.(2023)Zou, Phan, Chen, Campbell, Guo, Ren, Pan, Yin, Mazeika, Dombrowski, et~al.]{zou2023representation}
Zou, A., Phan, L., Chen, S., Campbell, J., Guo, P., Ren, R., Pan, A., Yin, X., Mazeika, M., Dombrowski, A.-K., et~al.
\newblock Representation engineering: A top-down approach to ai transparency.
\newblock \emph{arXiv preprint arXiv:2310.01405}, 2023.

\end{thebibliography}
\bibliographystyle{icml2026}

\newpage
\appendix
\crefalias{section}{appendix}
\crefname{appendix}{Appendix}{Appendices}
\Crefname{appendix}{Appendix}{Appendices}
\onecolumn

\section{Per-Layer Steering with Greedy Decoding}
\label{appendix:per-layer-steering-temp-0}

\begin{figure}[H] 
  \centering
  \begin{subfigure}{0.48\textwidth}
    \centering
    \includegraphics[width=\linewidth,page=1]{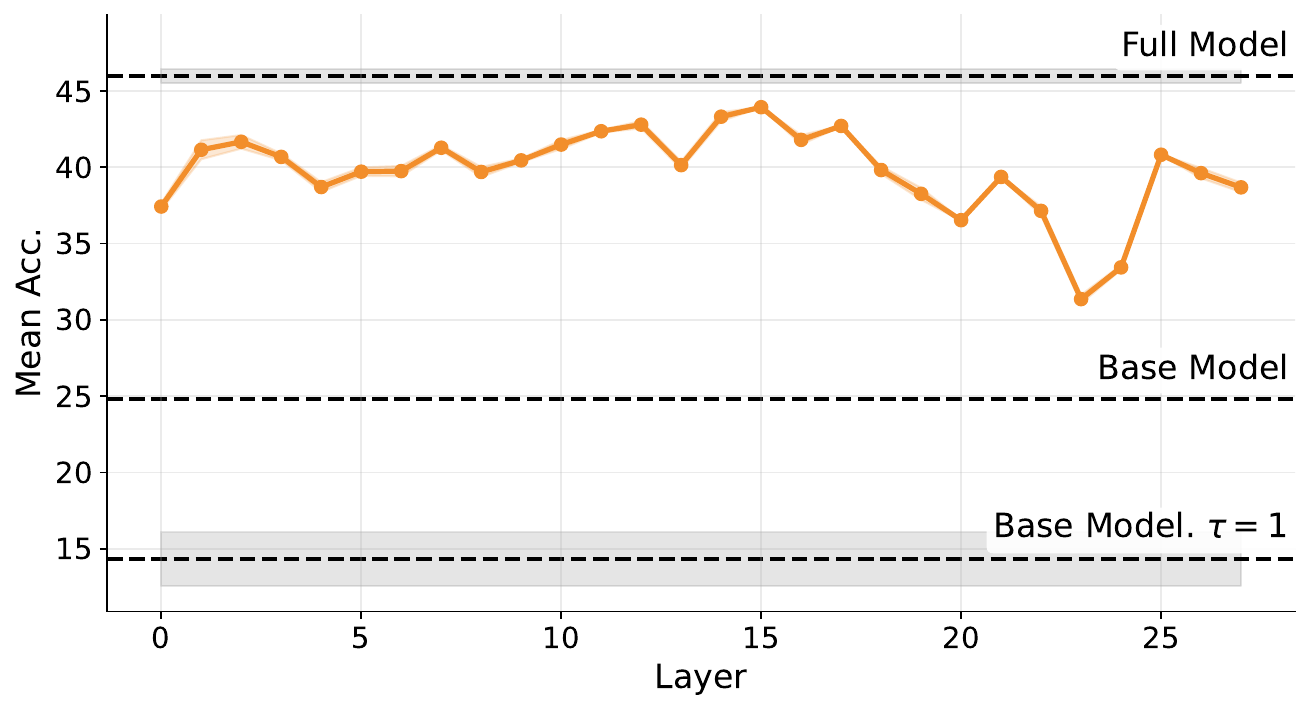}
    \caption*{Qwen2.5-Math-7B}
    \label{fig:per-layer-steering-qwen-temp-0}
  \end{subfigure}\hfill
  \begin{subfigure}{0.48\textwidth}
    \centering
    \includegraphics[width=\linewidth,page=1]{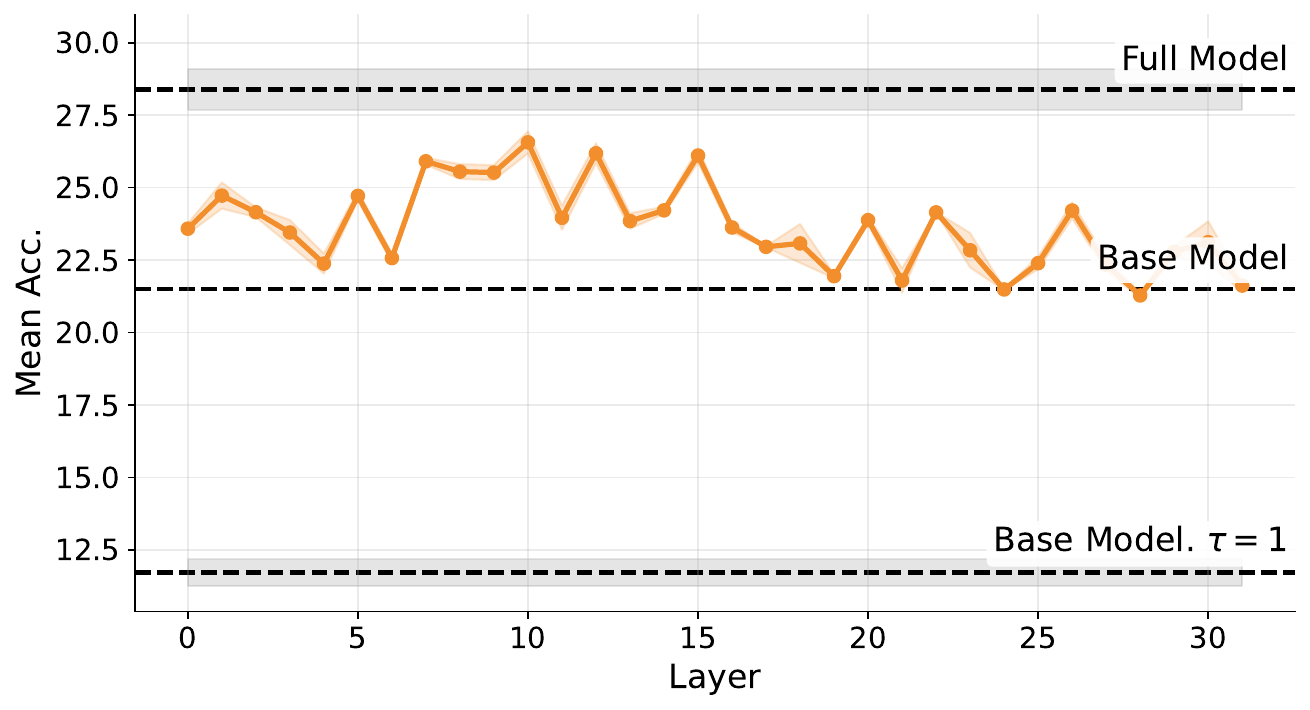}
    \caption*{Llama3.1-8B-It}
    \label{fig:per-layer-steering-llama-temp-0}
  \end{subfigure}
  \caption{\textbf{Single-layer steering with $\tau=0$.} We re-evaluated the single-layer steering vectors from \Cref{sec:svs_match} under greedy decoding.}
  \label{fig:per-layer-steering-temp-0}
\end{figure}

\section{Raw Benchmark Scores. Layers}
\label{appendix:raw_bench_scores_layers}

\begin{table*}[h]
\centering
\caption{Raw benchmark scores for Qwen2.5-Math-7B in \Cref{fig:per-layer-steering}.}
\label{tab:raw_bench_scores_layers_qwen_temp_1}
\small
\pgfplotstabletypeset[
  col sep=comma,
  string type,
  columns={Setup,AIME2025,AIME_2024,AMC_23,MATH500,MinervaMath,OlympiadBench,Avg},
  every head row/.style={before row=\toprule,after row=\midrule},
  every last row/.style={after row=\bottomrule},
  columns/Setup/.style={column name={Setup}, string type},
  columns/AIME2025/.style={column name={AIME25}, string type},
  columns/AIME_2024/.style={column name={AIME24}, string type},
  columns/AMC_23/.style={column name={AMC23}, string type},
  columns/MATH500/.style={column name={MATH500}, string type},
  columns/MinervaMath/.style={column name={MinervaMath}, string type},
  columns/OlympiadBench/.style={column name={OlympiadBench}, string type},
  columns/Avg/.style={column name={Avg.}, string type}
]{%
Setup,AIME2025,AIME_2024,AMC_23,MATH500,MinervaMath,OlympiadBench,Avg
Base Model,3.3 $\pm$ 0.0,16.7 $\pm$ 0.0,45.8 $\pm$ 0.0,52.2 $\pm$ 0.0,12.3 $\pm$ 0.0,18.6 $\pm$ 0.0,24.8 $\pm$ 0.0
Base Model. $\tau=1$,2.2 $\pm$ 1.6,10.0 $\pm$ 4.7,25.0 $\pm$ 8.2,37.7 $\pm$ 6.1,8.3 $\pm$ 1.5,10.2 $\pm$ 2.9,14.3 $\pm$ 1.7
Full-Tune,13.3 $\pm$ 2.7,30.0 $\pm$ 4.7,64.2 $\pm$ 3.1,79.6 $\pm$ 1.2,37.3 $\pm$ 0.8,41.7 $\pm$ 0.5,43.8 $\pm$ 0.4
Steering,17.8 $\pm$ 6.3,18.9 $\pm$ 1.6,63.3 $\pm$ 4.2,79.8 $\pm$ 0.3,35.8 $\pm$ 1.1,42.8 $\pm$ 1.1,42.9 $\pm$ 0.2
Layer-0,7.8 $\pm$ 1.6,12.2 $\pm$ 1.6,54.2 $\pm$ 4.2,69.5 $\pm$ 1.6,29.4 $\pm$ 1.0,33.7 $\pm$ 1.1,35.1 $\pm$ 0.5
Layer-1,5.6 $\pm$ 1.6,14.4 $\pm$ 6.3,55.8 $\pm$ 9.2,72.9 $\pm$ 2.1,27.7 $\pm$ 2.0,35.9 $\pm$ 0.4,35.8 $\pm$ 0.4
Layer-2,7.8 $\pm$ 1.6,21.1 $\pm$ 6.8,50.8 $\pm$ 4.7,70.4 $\pm$ 1.1,29.4 $\pm$ 2.3,34.3 $\pm$ 0.6,35.6 $\pm$ 0.6
Layer-3,6.7 $\pm$ 2.7,17.8 $\pm$ 1.6,50.0 $\pm$ 7.4,71.5 $\pm$ 1.0,30.1 $\pm$ 1.9,36.9 $\pm$ 1.1,36.4 $\pm$ 0.4
Layer-4,7.8 $\pm$ 1.6,23.3 $\pm$ 2.7,57.5 $\pm$ 4.1,71.5 $\pm$ 0.7,31.6 $\pm$ 0.8,37.1 $\pm$ 0.5,36.7 $\pm$ 0.1
Layer-5,12.2 $\pm$ 1.6,16.7 $\pm$ 2.7,58.3 $\pm$ 3.1,75.1 $\pm$ 0.8,28.6 $\pm$ 1.2,36.4 $\pm$ 1.0,36.9 $\pm$ 0.4
Layer-6,12.2 $\pm$ 4.2,17.8 $\pm$ 3.1,55.8 $\pm$ 3.1,72.7 $\pm$ 1.1,27.6 $\pm$ 1.4,36.3 $\pm$ 0.6,36.4 $\pm$ 0.4
Layer-7,11.1 $\pm$ 1.6,17.8 $\pm$ 3.1,55.0 $\pm$ 2.0,73.5 $\pm$ 0.7,30.3 $\pm$ 1.0,37.4 $\pm$ 0.7,37.3 $\pm$ 0.2
Layer-8,11.1 $\pm$ 4.2,20.0 $\pm$ 9.8,54.2 $\pm$ 1.2,74.1 $\pm$ 1.5,30.1 $\pm$ 2.9,35.4 $\pm$ 0.5,37.0 $\pm$ 0.7
Layer-9,8.9 $\pm$ 1.6,15.6 $\pm$ 1.6,55.8 $\pm$ 5.1,75.2 $\pm$ 0.6,28.7 $\pm$ 3.0,38.6 $\pm$ 0.3,37.6 $\pm$ 0.5
Layer-10,12.2 $\pm$ 1.6,15.6 $\pm$ 9.6,51.7 $\pm$ 2.4,72.5 $\pm$ 0.4,31.0 $\pm$ 1.9,36.5 $\pm$ 1.4,37.2 $\pm$ 0.1
Layer-11,8.9 $\pm$ 1.6,18.9 $\pm$ 5.7,55.0 $\pm$ 7.4,74.7 $\pm$ 0.7,30.4 $\pm$ 2.2,37.0 $\pm$ 1.3,37.4 $\pm$ 0.5
Layer-12,11.1 $\pm$ 5.7,18.9 $\pm$ 1.6,58.3 $\pm$ 4.2,75.5 $\pm$ 0.8,28.4 $\pm$ 2.2,38.0 $\pm$ 1.1,38.1 $\pm$ 0.6
Layer-13,15.6 $\pm$ 3.1,23.3 $\pm$ 4.7,51.7 $\pm$ 5.1,75.7 $\pm$ 0.8,29.4 $\pm$ 1.5,38.4 $\pm$ 0.9,38.4 $\pm$ 0.6
Layer-14,10.0 $\pm$ 4.7,23.3 $\pm$ 5.4,55.0 $\pm$ 3.5,76.2 $\pm$ 0.6,29.7 $\pm$ 1.5,38.7 $\pm$ 0.6,40.0 $\pm$ 0.3
Layer-15,13.3 $\pm$ 2.7,15.6 $\pm$ 4.2,61.7 $\pm$ 6.2,75.1 $\pm$ 0.2,30.9 $\pm$ 3.0,39.2 $\pm$ 0.8,39.2 $\pm$ 0.7
Layer-16,13.3 $\pm$ 2.7,17.8 $\pm$ 4.2,65.0 $\pm$ 4.1,76.6 $\pm$ 1.8,32.4 $\pm$ 1.3,40.3 $\pm$ 0.7,40.1 $\pm$ 0.6
Layer-17,8.9 $\pm$ 4.2,17.8 $\pm$ 5.7,56.7 $\pm$ 5.1,72.9 $\pm$ 0.7,27.7 $\pm$ 1.0,36.9 $\pm$ 1.4,36.6 $\pm$ 0.3
Layer-18,11.1 $\pm$ 1.6,11.1 $\pm$ 4.2,57.5 $\pm$ 3.5,74.7 $\pm$ 0.6,28.2 $\pm$ 2.0,39.8 $\pm$ 1.4,37.5 $\pm$ 0.5
Layer-19,14.4 $\pm$ 3.1,13.3 $\pm$ 0.0,59.2 $\pm$ 3.1,74.9 $\pm$ 0.9,33.0 $\pm$ 0.5,39.7 $\pm$ 0.7,38.4 $\pm$ 0.2
Layer-20,7.8 $\pm$ 3.1,15.6 $\pm$ 1.6,56.7 $\pm$ 5.1,74.0 $\pm$ 0.7,30.9 $\pm$ 1.8,37.0 $\pm$ 0.6,36.6 $\pm$ 0.2
Layer-21,6.7 $\pm$ 2.7,12.2 $\pm$ 1.6,61.7 $\pm$ 1.2,75.1 $\pm$ 0.9,30.3 $\pm$ 1.1,39.1 $\pm$ 0.7,37.5 $\pm$ 0.1
Layer-22,8.9 $\pm$ 1.6,16.7 $\pm$ 4.7,53.3 $\pm$ 3.1,72.3 $\pm$ 1.0,28.1 $\pm$ 0.9,35.6 $\pm$ 1.4,35.5 $\pm$ 0.2
Layer-23,5.6 $\pm$ 1.6,14.4 $\pm$ 6.3,35.8 $\pm$ 1.2,46.8 $\pm$ 5.0,13.6 $\pm$ 2.6,20.1 $\pm$ 5.4,20.9 $\pm$ 2.1
Layer-24,6.7 $\pm$ 2.7,12.2 $\pm$ 1.6,39.2 $\pm$ 2.4,47.7 $\pm$ 5.6,12.0 $\pm$ 2.6,20.4 $\pm$ 5.2,21.1 $\pm$ 2.2
Layer-25,7.8 $\pm$ 1.6,27.8 $\pm$ 1.6,60.8 $\pm$ 1.2,72.9 $\pm$ 0.8,31.6 $\pm$ 2.4,36.7 $\pm$ 0.7,38.2 $\pm$ 0.5
Layer-26,10.0 $\pm$ 2.7,15.6 $\pm$ 6.8,56.7 $\pm$ 7.2,72.9 $\pm$ 1.1,27.7 $\pm$ 4.1,37.1 $\pm$ 0.5,36.9 $\pm$ 0.6
Layer-27,5.6 $\pm$ 3.1,11.1 $\pm$ 1.6,49.2 $\pm$ 8.2,60.9 $\pm$ 0.4,20.5 $\pm$ 0.3,29.6 $\pm$ 1.3,29.4 $\pm$ 0.4
}
\end{table*}

\begin{table*}[h]
\centering
\caption{Raw benchmark scores for Llama3.1-8b-It in \Cref{fig:per-layer-steering}.}
\label{tab:raw_bench_scores_layers_llama_temp_1}
\small
\pgfplotstabletypeset[
  col sep=comma,
  string type,
  columns={Setup,AIME2025,AIME_2024,AMC_23,MATH500,MinervaMath,OlympiadBench,Avg},
  every head row/.style={before row=\toprule,after row=\midrule},
  every last row/.style={after row=\bottomrule},
  columns/Setup/.style={column name={Setup}, string type},
  columns/AIME2025/.style={column name={AIME25}, string type},
  columns/AIME_2024/.style={column name={AIME24}, string type},
  columns/AMC_23/.style={column name={AMC23}, string type},
  columns/MATH500/.style={column name={MATH500}, string type},
  columns/MinervaMath/.style={column name={MinervaMath}, string type},
  columns/OlympiadBench/.style={column name={OlympiadBench}, string type},
  columns/Avg/.style={column name={Avg.}, string type}
]{%
Setup,AIME2025,AIME_2024,AMC_23,MATH500,MinervaMath,OlympiadBench,Avg
Base Model,0.0 $\pm$ 0.0,10.0 $\pm$ 0.0,27.5 $\pm$ 0.0,52.5 $\pm$ 0.0,20.5 $\pm$ 0.0,18.6 $\pm$ 0.0,21.5 $\pm$ 0.0
Base Model. $\tau=1$,0.0 $\pm$ 0.0,0.0 $\pm$ 0.0,11.7 $\pm$ 2.4,34.6 $\pm$ 0.7,12.3 $\pm$ 1.5,9.6 $\pm$ 0.6,11.7 $\pm$ 0.5
Full-Tune,0.0 $\pm$ 0.0,8.9 $\pm$ 3.1,36.7 $\pm$ 1.2,57.6 $\pm$ 1.6,30.4 $\pm$ 2.3,22.4 $\pm$ 0.1,26.4 $\pm$ 0.8
Steering,0.0 $\pm$ 0.0,11.1 $\pm$ 5.7,35.0 $\pm$ 6.1,57.6 $\pm$ 1.9,29.9 $\pm$ 0.6,23.4 $\pm$ 0.5,25.8 $\pm$ 0.2
Layer-0,0.0 $\pm$ 0.0,10.0 $\pm$ 2.7,23.3 $\pm$ 6.6,50.7 $\pm$ 0.5,21.7 $\pm$ 0.9,19.1 $\pm$ 0.8,21.0 $\pm$ 0.4
Layer-1,1.1 $\pm$ 1.6,3.3 $\pm$ 2.7,30.8 $\pm$ 2.4,52.1 $\pm$ 0.9,24.5 $\pm$ 1.1,19.0 $\pm$ 1.3,22.0 $\pm$ 0.4
Layer-2,0.0 $\pm$ 0.0,10.0 $\pm$ 2.7,35.0 $\pm$ 2.0,53.1 $\pm$ 1.9,25.6 $\pm$ 1.5,18.7 $\pm$ 0.6,22.6 $\pm$ 0.4
Layer-3,0.0 $\pm$ 0.0,6.7 $\pm$ 2.7,30.0 $\pm$ 6.1,53.9 $\pm$ 0.4,28.6 $\pm$ 0.5,18.8 $\pm$ 0.6,22.6 $\pm$ 0.1
Layer-4,0.0 $\pm$ 0.0,7.8 $\pm$ 1.6,23.3 $\pm$ 3.1,50.8 $\pm$ 0.5,26.2 $\pm$ 1.0,19.0 $\pm$ 0.9,21.6 $\pm$ 0.2
Layer-5,0.0 $\pm$ 0.0,6.7 $\pm$ 2.7,29.2 $\pm$ 4.7,52.7 $\pm$ 1.0,26.2 $\pm$ 2.3,19.3 $\pm$ 0.3,22.3 $\pm$ 0.3
Layer-6,1.1 $\pm$ 1.6,4.4 $\pm$ 3.1,30.0 $\pm$ 5.4,50.8 $\pm$ 0.7,25.6 $\pm$ 1.2,19.8 $\pm$ 1.1,22.0 $\pm$ 0.1
Layer-7,0.0 $\pm$ 0.0,4.4 $\pm$ 3.1,27.5 $\pm$ 7.1,53.3 $\pm$ 0.5,27.5 $\pm$ 1.5,20.2 $\pm$ 0.2,22.6 $\pm$ 0.2
Layer-8,0.0 $\pm$ 0.0,7.8 $\pm$ 4.2,30.8 $\pm$ 5.1,53.3 $\pm$ 1.4,27.5 $\pm$ 1.4,19.7 $\pm$ 0.4,23.1 $\pm$ 0.2
Layer-9,1.1 $\pm$ 1.6,5.6 $\pm$ 1.6,30.8 $\pm$ 3.1,56.9 $\pm$ 0.7,27.7 $\pm$ 1.0,21.9 $\pm$ 0.3,24.7 $\pm$ 0.1
Layer-10,0.0 $\pm$ 0.0,4.4 $\pm$ 4.2,33.3 $\pm$ 4.7,55.5 $\pm$ 1.2,27.7 $\pm$ 1.0,21.3 $\pm$ 0.9,23.4 $\pm$ 0.1
Layer-11,1.1 $\pm$ 1.6,4.4 $\pm$ 1.6,33.3 $\pm$ 2.4,55.1 $\pm$ 1.4,25.9 $\pm$ 0.6,20.6 $\pm$ 0.8,24.2 $\pm$ 0.1
Layer-12,1.1 $\pm$ 1.6,7.8 $\pm$ 3.1,32.5 $\pm$ 2.0,54.3 $\pm$ 1.3,29.0 $\pm$ 1.4,22.3 $\pm$ 0.7,24.5 $\pm$ 0.4
Layer-13,1.1 $\pm$ 1.6,7.8 $\pm$ 4.2,24.2 $\pm$ 1.2,54.7 $\pm$ 1.9,27.6 $\pm$ 2.9,20.1 $\pm$ 0.7,23.1 $\pm$ 0.1
Layer-14,0.0 $\pm$ 0.0,11.1 $\pm$ 4.2,31.7 $\pm$ 1.2,55.1 $\pm$ 0.4,28.4 $\pm$ 0.2,20.9 $\pm$ 0.1,24.0 $\pm$ 0.1
Layer-15,1.1 $\pm$ 1.6,4.4 $\pm$ 1.6,27.5 $\pm$ 2.0,54.0 $\pm$ 1.1,27.3 $\pm$ 1.4,21.8 $\pm$ 1.0,23.5 $\pm$ 0.4
Layer-16,0.0 $\pm$ 0.0,7.8 $\pm$ 3.1,25.8 $\pm$ 3.1,52.9 $\pm$ 0.1,26.8 $\pm$ 0.6,20.6 $\pm$ 0.3,22.7 $\pm$ 0.1
Layer-17,0.0 $\pm$ 0.0,5.6 $\pm$ 1.6,26.7 $\pm$ 1.2,52.1 $\pm$ 0.9,25.7 $\pm$ 0.5,18.6 $\pm$ 0.7,22.0 $\pm$ 0.2
Layer-18,0.0 $\pm$ 0.0,10.0 $\pm$ 5.4,18.3 $\pm$ 6.6,51.5 $\pm$ 1.2,25.5 $\pm$ 1.4,19.3 $\pm$ 1.2,21.6 $\pm$ 0.5
Layer-19,1.1 $\pm$ 1.6,6.7 $\pm$ 2.7,30.0 $\pm$ 3.5,52.9 $\pm$ 1.1,25.7 $\pm$ 2.0,19.3 $\pm$ 0.7,22.3 $\pm$ 0.6
Layer-20,0.0 $\pm$ 0.0,6.7 $\pm$ 2.7,22.5 $\pm$ 4.1,52.9 $\pm$ 1.1,27.1 $\pm$ 1.7,19.3 $\pm$ 1.1,21.9 $\pm$ 0.1
Layer-21,0.0 $\pm$ 0.0,4.4 $\pm$ 1.6,25.8 $\pm$ 2.4,51.7 $\pm$ 1.0,27.0 $\pm$ 0.2,18.6 $\pm$ 0.4,21.2 $\pm$ 0.3
Layer-22,1.1 $\pm$ 1.6,7.8 $\pm$ 5.7,22.5 $\pm$ 2.0,54.9 $\pm$ 0.4,25.4 $\pm$ 2.5,18.1 $\pm$ 0.6,22.5 $\pm$ 0.6
Layer-23,1.1 $\pm$ 1.6,6.7 $\pm$ 2.7,25.8 $\pm$ 1.2,52.1 $\pm$ 0.2,25.7 $\pm$ 0.8,19.0 $\pm$ 1.5,21.1 $\pm$ 0.2
Layer-24,2.2 $\pm$ 1.6,5.6 $\pm$ 3.1,22.5 $\pm$ 7.4,50.6 $\pm$ 2.2,27.0 $\pm$ 2.5,18.8 $\pm$ 1.0,21.4 $\pm$ 0.6
Layer-25,0.0 $\pm$ 0.0,6.7 $\pm$ 2.7,27.5 $\pm$ 2.0,52.1 $\pm$ 0.5,26.6 $\pm$ 1.1,20.3 $\pm$ 1.5,21.9 $\pm$ 0.4
Layer-26,2.2 $\pm$ 1.6,5.6 $\pm$ 1.6,27.5 $\pm$ 2.0,50.5 $\pm$ 1.4,28.1 $\pm$ 1.8,17.8 $\pm$ 1.3,21.6 $\pm$ 0.5
Layer-27,1.1 $\pm$ 1.6,4.4 $\pm$ 1.6,24.2 $\pm$ 4.2,50.5 $\pm$ 0.1,28.1 $\pm$ 0.9,18.2 $\pm$ 0.2,21.5 $\pm$ 0.2
Layer-28,1.1 $\pm$ 1.6,5.6 $\pm$ 1.6,20.0 $\pm$ 3.5,50.5 $\pm$ 1.7,25.5 $\pm$ 1.1,17.4 $\pm$ 0.3,20.9 $\pm$ 0.1
Layer-29,0.0 $\pm$ 0.0,6.7 $\pm$ 2.7,30.0 $\pm$ 3.5,50.5 $\pm$ 0.8,25.7 $\pm$ 2.1,16.9 $\pm$ 0.6,21.1 $\pm$ 0.6
Layer-30,0.0 $\pm$ 0.0,7.8 $\pm$ 4.2,24.2 $\pm$ 2.4,50.4 $\pm$ 1.0,25.0 $\pm$ 0.8,15.8 $\pm$ 1.4,20.1 $\pm$ 0.5
Layer-31,1.1 $\pm$ 1.6,1.1 $\pm$ 1.6,18.3 $\pm$ 1.2,39.0 $\pm$ 0.7,18.1 $\pm$ 2.5,11.7 $\pm$ 0.4,14.7 $\pm$ 0.5
}
\end{table*}

\begin{table*}[h]
\centering
\caption{Raw benchmark scores for Qwen2.5-Math-7B in \Cref{fig:per-layer-steering-temp-0} (greedy decoding).}
\label{tab:raw_bench_scores_layers_qwen_temp_0}
\small
\pgfplotstabletypeset[
  col sep=comma,
  string type,
  columns={Setup,AIME2025,AIME_2024,AMC_23,MATH500,MinervaMath,OlympiadBench,Avg},
  every head row/.style={before row=\toprule,after row=\midrule},
  every last row/.style={after row=\bottomrule},
  columns/Setup/.style={column name={Setup}, string type},
  columns/AIME2025/.style={column name={AIME25}, string type},
  columns/AIME_2024/.style={column name={AIME24}, string type},
  columns/AMC_23/.style={column name={AMC23}, string type},
  columns/MATH500/.style={column name={MATH500}, string type},
  columns/MinervaMath/.style={column name={MinervaMath}, string type},
  columns/OlympiadBench/.style={column name={OlympiadBench}, string type},
  columns/Avg/.style={column name={Avg.}, string type}
]{%
Setup,AIME2025,AIME_2024,AMC_23,MATH500,MinervaMath,OlympiadBench,Avg
Base Model,3.3 $\pm$ 0.0,16.7 $\pm$ 0.0,45.8 $\pm$ 0.0,52.2 $\pm$ 0.0,12.3 $\pm$ 0.0,18.6 $\pm$ 0.0,24.8 $\pm$ 0.0
Base Model. $\tau=1$,2.2 $\pm$ 1.6,10.0 $\pm$ 4.7,25.0 $\pm$ 8.2,37.7 $\pm$ 6.1,8.3 $\pm$ 1.5,10.2 $\pm$ 2.9,14.3 $\pm$ 1.7
Full-Tune,18.9 $\pm$ 1.6,31.1 $\pm$ 1.6,67.5 $\pm$ 0.0,80.1 $\pm$ 0.1,34.9 $\pm$ 0.0,43.2 $\pm$ 0.2,46.0 $\pm$ 0.5
Steering,16.7 $\pm$ 0.0,24.4 $\pm$ 1.6,59.2 $\pm$ 1.2,79.1 $\pm$ 0.2,33.9 $\pm$ 0.3,42.0 $\pm$ 0.1,42.6 $\pm$ 0.4
Layer-0,8.9 $\pm$ 1.6,13.3 $\pm$ 0.0,57.5 $\pm$ 0.0,75.3 $\pm$ 0.1,29.7 $\pm$ 0.2,39.9 $\pm$ 0.2,37.4 $\pm$ 0.2
Layer-1,13.3 $\pm$ 2.7,20.0 $\pm$ 0.0,68.3 $\pm$ 1.2,77.5 $\pm$ 0.2,28.4 $\pm$ 0.2,39.3 $\pm$ 0.1,41.1 $\pm$ 0.6
Layer-2,16.7 $\pm$ 2.7,20.0 $\pm$ 0.0,65.0 $\pm$ 0.0,77.9 $\pm$ 0.2,30.8 $\pm$ 0.2,39.7 $\pm$ 0.1,41.7 $\pm$ 0.4
Layer-3,10.0 $\pm$ 0.0,26.7 $\pm$ 0.0,60.8 $\pm$ 1.2,76.2 $\pm$ 0.5,29.8 $\pm$ 0.0,40.6 $\pm$ 0.1,40.7 $\pm$ 0.2
Layer-4,11.1 $\pm$ 1.6,10.0 $\pm$ 0.0,65.0 $\pm$ 0.0,75.3 $\pm$ 0.5,30.9 $\pm$ 0.3,39.9 $\pm$ 0.3,38.7 $\pm$ 0.3
Layer-5,15.6 $\pm$ 1.6,20.0 $\pm$ 0.0,57.5 $\pm$ 0.0,76.7 $\pm$ 0.1,29.2 $\pm$ 0.2,39.3 $\pm$ 0.3,39.7 $\pm$ 0.3
Layer-6,16.7 $\pm$ 0.0,18.9 $\pm$ 1.6,58.3 $\pm$ 1.2,78.6 $\pm$ 0.3,27.3 $\pm$ 0.2,38.7 $\pm$ 0.1,39.7 $\pm$ 0.3
Layer-7,13.3 $\pm$ 0.0,23.3 $\pm$ 0.0,60.8 $\pm$ 1.2,79.2 $\pm$ 0.2,29.0 $\pm$ 0.0,41.9 $\pm$ 0.1,41.3 $\pm$ 0.2
Layer-8,7.8 $\pm$ 1.6,16.7 $\pm$ 0.0,67.5 $\pm$ 0.0,78.9 $\pm$ 0.1,27.6 $\pm$ 0.3,39.7 $\pm$ 0.1,39.7 $\pm$ 0.3
Layer-9,13.3 $\pm$ 0.0,23.3 $\pm$ 0.0,55.0 $\pm$ 0.0,79.3 $\pm$ 0.2,30.4 $\pm$ 0.3,41.4 $\pm$ 0.2,40.5 $\pm$ 0.1
Layer-10,16.7 $\pm$ 0.0,26.7 $\pm$ 0.0,57.5 $\pm$ 2.0,77.5 $\pm$ 0.1,31.2 $\pm$ 0.3,39.3 $\pm$ 0.1,41.5 $\pm$ 0.3
Layer-11,12.2 $\pm$ 1.6,30.0 $\pm$ 0.0,58.3 $\pm$ 1.2,78.5 $\pm$ 0.2,33.5 $\pm$ 0.0,41.7 $\pm$ 0.3,42.4 $\pm$ 0.1
Layer-12,16.7 $\pm$ 0.0,21.1 $\pm$ 1.6,65.8 $\pm$ 1.2,77.8 $\pm$ 0.0,33.5 $\pm$ 0.0,41.9 $\pm$ 0.4,42.8 $\pm$ 0.2
Layer-13,16.7 $\pm$ 0.0,16.7 $\pm$ 0.0,55.8 $\pm$ 1.2,79.3 $\pm$ 0.3,31.1 $\pm$ 0.2,41.3 $\pm$ 0.1,40.1 $\pm$ 0.3
Layer-14,15.6 $\pm$ 1.6,26.7 $\pm$ 0.0,62.5 $\pm$ 0.0,79.2 $\pm$ 0.2,31.6 $\pm$ 0.0,44.3 $\pm$ 0.3,43.3 $\pm$ 0.3
Layer-15,16.7 $\pm$ 0.0,23.3 $\pm$ 0.0,70.0 $\pm$ 0.0,78.4 $\pm$ 0.2,32.8 $\pm$ 0.2,42.4 $\pm$ 0.1,43.9 $\pm$ 0.0
Layer-16,8.9 $\pm$ 1.6,26.7 $\pm$ 0.0,60.0 $\pm$ 0.0,79.1 $\pm$ 0.1,36.4 $\pm$ 0.0,39.8 $\pm$ 0.3,41.8 $\pm$ 0.3
Layer-17,13.3 $\pm$ 0.0,30.0 $\pm$ 0.0,57.5 $\pm$ 0.0,78.1 $\pm$ 0.1,34.6 $\pm$ 0.0,42.7 $\pm$ 0.1,42.7 $\pm$ 0.0
Layer-18,12.2 $\pm$ 1.6,13.3 $\pm$ 0.0,65.0 $\pm$ 0.0,78.6 $\pm$ 0.0,29.4 $\pm$ 0.0,40.3 $\pm$ 0.1,39.8 $\pm$ 0.3
Layer-19,16.7 $\pm$ 0.0,13.3 $\pm$ 0.0,49.2 $\pm$ 2.4,77.8 $\pm$ 0.2,30.5 $\pm$ 0.3,42.1 $\pm$ 0.1,38.3 $\pm$ 0.4
Layer-20,0.0 $\pm$ 0.0,13.3 $\pm$ 0.0,62.5 $\pm$ 0.0,75.9 $\pm$ 0.1,30.1 $\pm$ 0.0,37.3 $\pm$ 0.1,36.5 $\pm$ 0.0
Layer-21,10.0 $\pm$ 0.0,13.3 $\pm$ 0.0,62.5 $\pm$ 0.0,78.5 $\pm$ 0.1,32.7 $\pm$ 0.0,39.1 $\pm$ 0.1,39.4 $\pm$ 0.0
Layer-22,8.9 $\pm$ 3.1,13.3 $\pm$ 0.0,59.2 $\pm$ 1.2,75.5 $\pm$ 0.2,27.3 $\pm$ 0.2,38.7 $\pm$ 0.1,37.1 $\pm$ 0.3
Layer-23,4.4 $\pm$ 1.6,23.3 $\pm$ 0.0,57.5 $\pm$ 0.0,60.4 $\pm$ 0.0,17.4 $\pm$ 0.2,25.1 $\pm$ 0.1,31.4 $\pm$ 0.3
Layer-24,16.7 $\pm$ 0.0,20.0 $\pm$ 0.0,49.2 $\pm$ 1.2,65.1 $\pm$ 0.1,18.6 $\pm$ 0.5,31.1 $\pm$ 0.3,33.4 $\pm$ 0.2
Layer-25,16.7 $\pm$ 0.0,23.3 $\pm$ 0.0,60.0 $\pm$ 0.0,77.0 $\pm$ 0.0,30.4 $\pm$ 0.2,37.5 $\pm$ 0.1,40.8 $\pm$ 0.0
Layer-26,6.7 $\pm$ 0.0,30.0 $\pm$ 0.0,60.0 $\pm$ 2.0,74.5 $\pm$ 0.2,27.6 $\pm$ 0.0,38.9 $\pm$ 0.1,39.6 $\pm$ 0.3
Layer-27,13.3 $\pm$ 0.0,17.8 $\pm$ 1.6,55.8 $\pm$ 1.2,72.5 $\pm$ 0.1,34.1 $\pm$ 0.3,38.6 $\pm$ 0.4,38.7 $\pm$ 0.3
}
\end{table*}

\begin{table*}[h]
\centering
\caption{Raw benchmark scores for Llama3.1-8b-It in \Cref{fig:per-layer-steering-temp-0} (greedy decoding).}
\label{tab:raw_bench_scores_layers_llama_temp_0}
\small
\pgfplotstabletypeset[
  col sep=comma,
  string type,
  columns={Setup,AIME2025,AIME_2024,AMC_23,MATH500,MinervaMath,OlympiadBench,Avg},
  every head row/.style={before row=\toprule,after row=\midrule},
  every last row/.style={after row=\bottomrule},
  columns/Setup/.style={column name={Setup}, string type},
  columns/AIME2025/.style={column name={AIME25}, string type},
  columns/AIME_2024/.style={column name={AIME24}, string type},
  columns/AMC_23/.style={column name={AMC23}, string type},
  columns/MATH500/.style={column name={MATH500}, string type},
  columns/MinervaMath/.style={column name={MinervaMath}, string type},
  columns/OlympiadBench/.style={column name={OlympiadBench}, string type},
  columns/Avg/.style={column name={Avg.}, string type}
]{%
Setup,AIME2025,AIME_2024,AMC_23,MATH500,MinervaMath,OlympiadBench,Avg
Base Model,0.0 $\pm$ 0.0,10.0 $\pm$ 0.0,27.5 $\pm$ 0.0,52.5 $\pm$ 0.0,20.5 $\pm$ 0.0,18.6 $\pm$ 0.0,21.5 $\pm$ 0.0
Base Model. $\tau=1$,0.0 $\pm$ 0.0,0.0 $\pm$ 0.0,11.7 $\pm$ 2.4,34.6 $\pm$ 0.7,12.3 $\pm$ 1.5,9.6 $\pm$ 0.6,11.7 $\pm$ 0.5
Full-Tune,3.3 $\pm$ 0.0,14.4 $\pm$ 3.1,41.7 $\pm$ 2.4,57.7 $\pm$ 0.2,30.9 $\pm$ 0.6,22.3 $\pm$ 0.3,28.4 $\pm$ 0.7
Steering,0.0 $\pm$ 0.0,11.1 $\pm$ 5.7,35.0 $\pm$ 6.1,57.6 $\pm$ 1.9,29.9 $\pm$ 0.6,23.4 $\pm$ 0.5,25.8 $\pm$ 0.2
Layer-0,3.3 $\pm$ 0.0,6.7 $\pm$ 0.0,33.3 $\pm$ 1.2,53.7 $\pm$ 0.2,24.9 $\pm$ 0.2,19.6 $\pm$ 0.0,23.6 $\pm$ 0.2
Layer-1,6.7 $\pm$ 0.0,8.9 $\pm$ 1.6,35.8 $\pm$ 1.2,50.8 $\pm$ 0.0,27.5 $\pm$ 0.2,18.7 $\pm$ 0.2,24.7 $\pm$ 0.5
Layer-2,0.0 $\pm$ 0.0,3.3 $\pm$ 0.0,44.2 $\pm$ 1.2,51.6 $\pm$ 0.0,25.9 $\pm$ 0.2,20.0 $\pm$ 0.3,24.2 $\pm$ 0.2
Layer-3,0.0 $\pm$ 0.0,5.6 $\pm$ 1.6,32.5 $\pm$ 2.0,53.9 $\pm$ 0.1,28.9 $\pm$ 0.6,19.8 $\pm$ 0.1,23.5 $\pm$ 0.4
Layer-4,0.0 $\pm$ 0.0,6.7 $\pm$ 0.0,29.2 $\pm$ 1.2,53.1 $\pm$ 0.5,27.1 $\pm$ 0.2,18.2 $\pm$ 0.2,22.4 $\pm$ 0.3
Layer-5,0.0 $\pm$ 0.0,10.0 $\pm$ 0.0,36.7 $\pm$ 1.2,53.5 $\pm$ 0.1,27.2 $\pm$ 0.3,20.9 $\pm$ 0.1,24.7 $\pm$ 0.2
Layer-6,0.0 $\pm$ 0.0,6.7 $\pm$ 0.0,28.3 $\pm$ 1.2,54.4 $\pm$ 0.2,27.7 $\pm$ 0.2,18.3 $\pm$ 0.3,22.6 $\pm$ 0.2
Layer-7,6.7 $\pm$ 0.0,6.7 $\pm$ 0.0,37.5 $\pm$ 0.0,54.3 $\pm$ 0.3,28.1 $\pm$ 0.2,22.2 $\pm$ 0.2,25.9 $\pm$ 0.1
Layer-8,3.3 $\pm$ 0.0,3.3 $\pm$ 0.0,45.8 $\pm$ 1.2,50.7 $\pm$ 0.2,28.1 $\pm$ 0.2,22.0 $\pm$ 0.3,25.6 $\pm$ 0.3
Layer-9,3.3 $\pm$ 0.0,15.6 $\pm$ 1.6,30.0 $\pm$ 0.0,54.7 $\pm$ 0.2,27.9 $\pm$ 0.3,21.6 $\pm$ 0.2,25.5 $\pm$ 0.3
Layer-10,0.0 $\pm$ 0.0,13.3 $\pm$ 2.7,40.0 $\pm$ 0.0,56.8 $\pm$ 0.0,28.3 $\pm$ 0.3,20.9 $\pm$ 0.3,26.6 $\pm$ 0.4
Layer-11,0.0 $\pm$ 0.0,11.1 $\pm$ 1.6,30.0 $\pm$ 2.0,52.7 $\pm$ 0.1,29.0 $\pm$ 0.0,20.9 $\pm$ 0.1,24.0 $\pm$ 0.4
Layer-12,3.3 $\pm$ 0.0,4.4 $\pm$ 1.6,39.2 $\pm$ 1.2,56.7 $\pm$ 0.2,29.5 $\pm$ 0.3,23.9 $\pm$ 0.3,26.2 $\pm$ 0.4
Layer-13,3.3 $\pm$ 0.0,7.8 $\pm$ 1.6,25.0 $\pm$ 0.0,56.6 $\pm$ 0.0,29.3 $\pm$ 0.2,21.1 $\pm$ 0.2,23.8 $\pm$ 0.3
Layer-14,0.0 $\pm$ 0.0,10.0 $\pm$ 0.0,30.0 $\pm$ 0.0,55.4 $\pm$ 0.5,27.6 $\pm$ 0.0,22.3 $\pm$ 0.3,24.2 $\pm$ 0.1
Layer-15,6.7 $\pm$ 0.0,6.7 $\pm$ 0.0,38.3 $\pm$ 1.2,54.5 $\pm$ 0.2,29.2 $\pm$ 0.3,21.3 $\pm$ 0.1,26.1 $\pm$ 0.2
Layer-16,0.0 $\pm$ 0.0,6.7 $\pm$ 0.0,34.2 $\pm$ 1.2,52.5 $\pm$ 0.5,29.5 $\pm$ 0.2,18.9 $\pm$ 0.1,23.6 $\pm$ 0.2
Layer-17,0.0 $\pm$ 0.0,3.3 $\pm$ 0.0,37.5 $\pm$ 0.0,51.4 $\pm$ 0.3,26.3 $\pm$ 0.2,19.2 $\pm$ 0.3,23.0 $\pm$ 0.0
Layer-18,3.3 $\pm$ 0.0,7.8 $\pm$ 3.1,29.2 $\pm$ 1.2,51.1 $\pm$ 0.2,27.2 $\pm$ 0.3,19.9 $\pm$ 0.1,23.1 $\pm$ 0.7
Layer-19,0.0 $\pm$ 0.0,3.3 $\pm$ 0.0,32.5 $\pm$ 0.0,51.6 $\pm$ 0.3,24.4 $\pm$ 0.2,19.9 $\pm$ 0.2,21.9 $\pm$ 0.1
Layer-20,3.3 $\pm$ 0.0,6.7 $\pm$ 0.0,34.2 $\pm$ 1.2,54.6 $\pm$ 0.3,26.7 $\pm$ 0.2,17.8 $\pm$ 0.3,23.9 $\pm$ 0.1
Layer-21,3.3 $\pm$ 0.0,3.3 $\pm$ 0.0,25.8 $\pm$ 2.4,54.4 $\pm$ 0.2,27.3 $\pm$ 0.2,16.5 $\pm$ 0.1,21.8 $\pm$ 0.4
Layer-22,0.0 $\pm$ 0.0,10.0 $\pm$ 0.0,32.5 $\pm$ 0.0,53.0 $\pm$ 0.3,27.0 $\pm$ 0.2,22.4 $\pm$ 0.1,24.1 $\pm$ 0.0
Layer-23,0.0 $\pm$ 0.0,7.8 $\pm$ 1.6,30.0 $\pm$ 2.0,51.9 $\pm$ 0.3,29.9 $\pm$ 0.5,17.5 $\pm$ 0.2,22.8 $\pm$ 0.6
Layer-24,0.0 $\pm$ 0.0,3.3 $\pm$ 0.0,30.0 $\pm$ 0.0,51.5 $\pm$ 0.1,27.2 $\pm$ 0.0,16.8 $\pm$ 0.1,21.5 $\pm$ 0.0
Layer-25,0.0 $\pm$ 0.0,10.0 $\pm$ 0.0,29.2 $\pm$ 1.2,52.3 $\pm$ 0.2,24.3 $\pm$ 0.0,18.7 $\pm$ 0.1,22.4 $\pm$ 0.2
Layer-26,0.0 $\pm$ 0.0,4.4 $\pm$ 1.6,42.5 $\pm$ 0.0,52.8 $\pm$ 0.0,28.2 $\pm$ 0.2,17.3 $\pm$ 0.0,24.2 $\pm$ 0.3
Layer-27,0.0 $\pm$ 0.0,3.3 $\pm$ 0.0,34.2 $\pm$ 1.2,51.5 $\pm$ 0.2,27.1 $\pm$ 0.2,18.0 $\pm$ 0.1,22.3 $\pm$ 0.2
Layer-28,0.0 $\pm$ 0.0,4.4 $\pm$ 1.6,29.2 $\pm$ 1.2,50.3 $\pm$ 0.1,27.2 $\pm$ 0.3,16.6 $\pm$ 0.1,21.3 $\pm$ 0.1
Layer-29,0.0 $\pm$ 0.0,3.3 $\pm$ 0.0,36.7 $\pm$ 1.2,52.9 $\pm$ 0.1,28.1 $\pm$ 0.2,15.8 $\pm$ 0.1,22.8 $\pm$ 0.2
Layer-30,0.0 $\pm$ 0.0,6.7 $\pm$ 0.0,36.7 $\pm$ 4.7,52.3 $\pm$ 0.1,24.9 $\pm$ 0.2,18.2 $\pm$ 0.2,23.1 $\pm$ 0.7
Layer-31,0.0 $\pm$ 0.0,10.0 $\pm$ 0.0,28.3 $\pm$ 1.2,50.5 $\pm$ 0.3,25.1 $\pm$ 0.2,15.8 $\pm$ 0.1,21.6 $\pm$ 0.2
}
\end{table*}

\FloatBarrier
\section{Ineffective Layers in Qwen2.5-Math-7B}
\label{appendix:bad_layers_qwen}

\begin{figure}[h] 

    \centering
    \includegraphics[width=\linewidth,page=1]{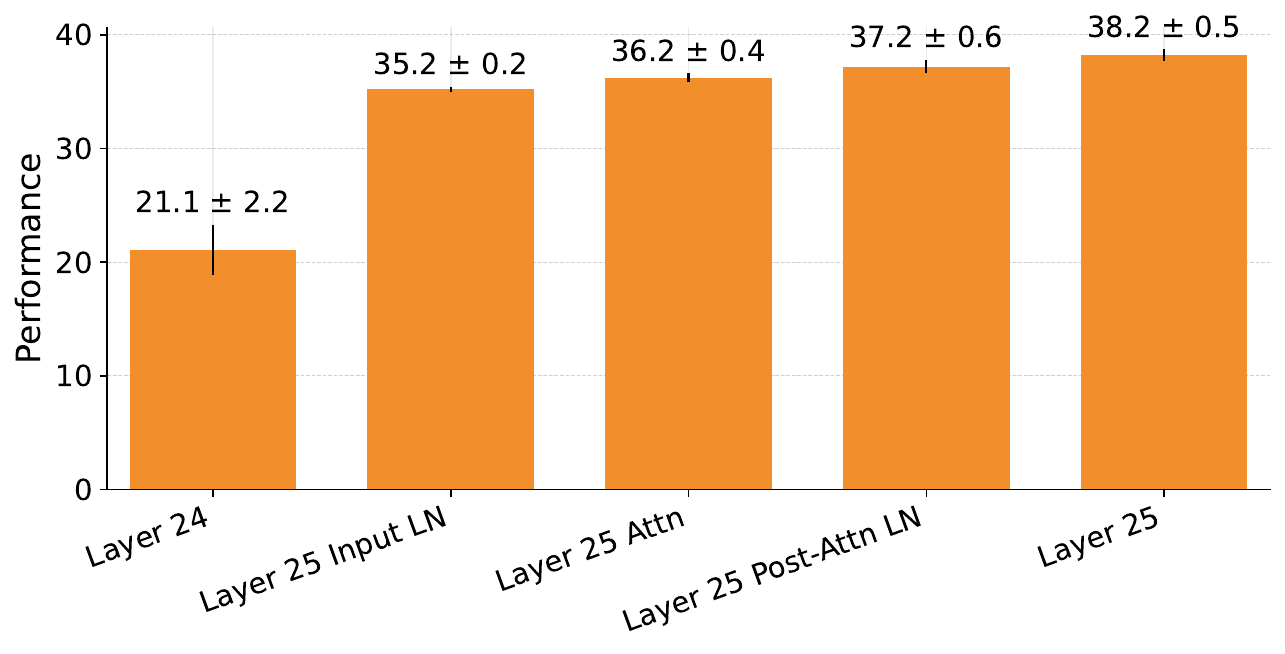}

  \caption{\textbf{Steering of a specific component's output.} We trained steering vectors on intermediate component outputs between layers 24 (where performance dropped) and 25 (where it recovered). Steering after the layer-25 input LayerNorm largely restored performance, indicating that the drop was mitigated once the intervention was carried past this normalization step.}

    \label{fig:bad-layers-qwen}
\end{figure}

As noted in \Cref{sec:svs_match}, single-layer steering on layers 23 and 24 underperforms their neighbors. To pinpoint where this loss arises, we trained vectors inserted immediately after each subcomponent between the layer-24 MLP and the layer-25 MLP. \Cref{fig:bad-layers-qwen} shows that placing \(s_{24}\) \emph{after} the input LayerNorm of layer 25 closes the gap with \(s_{25}\). Thus the input LayerNorm is the problematic step -- passing through it limits the effect of the steering vector.

\section{Steering Vector Persistence. Raw Numbers}
\label{appendix:persistence_raw}

\begin{table}[H]
  \centering
  \caption{\textbf{Qwen2.5-Math-7B.} Raw scores for the plots in \Cref{fig:persistence_both} (top).}
  \label{tab:qwen_persistence_raw}
  \begingroup
    \setlength{\tabcolsep}{4pt}
    \renewcommand{\arraystretch}{1.0}

    \begin{subtable}[t]{\textwidth}
      \centering
      \caption*{Diff-Diff CosSim}
      \begin{adjustbox}{max totalsize={\linewidth}{\textheight},center}
        \begin{filecontents*}{tables/qwen_diff_matrix.csv}
i,0,1,2,3,4,5,6,7,8,9,10,11,12,13,14,15,16,17,18,19,20,21,22,23,24,25,26,27
0,1.00,0.96,0.91,0.84,0.80,0.75,0.67,0.59,0.54,0.54,0.52,0.49,0.47,0.45,0.43,0.42,0.40,0.37,0.37,0.37,0.35,0.34,0.34,0.33,0.33,0.31,0.37,0.57
1,,1.00,0.95,0.85,0.80,0.74,0.65,0.58,0.51,0.51,0.49,0.45,0.43,0.42,0.39,0.38,0.36,0.34,0.33,0.33,0.31,0.32,0.32,0.31,0.31,0.29,0.29,0.41
2,,,1.00,0.96,0.91,0.86,0.78,0.70,0.64,0.64,0.61,0.57,0.54,0.52,0.50,0.48,0.45,0.42,0.42,0.41,0.39,0.37,0.37,0.37,0.36,0.34,0.38,0.55
3,,,,1.00,0.93,0.87,0.78,0.69,0.64,0.63,0.61,0.57,0.55,0.53,0.51,0.49,0.47,0.44,0.43,0.41,0.39,0.37,0.35,0.34,0.32,0.30,0.32,0.57
4,,,,,1.00,0.92,0.82,0.73,0.67,0.67,0.64,0.60,0.58,0.56,0.53,0.51,0.49,0.46,0.45,0.44,0.41,0.39,0.37,0.36,0.34,0.32,0.34,0.61
5,,,,,,1.00,0.88,0.77,0.71,0.69,0.66,0.62,0.59,0.56,0.53,0.51,0.49,0.46,0.44,0.43,0.40,0.38,0.37,0.36,0.35,0.32,0.36,0.62
6,,,,,,,1.00,0.82,0.72,0.70,0.65,0.61,0.58,0.55,0.52,0.50,0.47,0.44,0.44,0.42,0.39,0.38,0.36,0.35,0.33,0.32,0.33,0.59
7,,,,,,,,1.00,0.85,0.81,0.74,0.67,0.63,0.60,0.56,0.54,0.51,0.47,0.46,0.45,0.41,0.39,0.38,0.37,0.35,0.33,0.36,0.64
8,,,,,,,,,1.00,0.87,0.77,0.69,0.64,0.60,0.56,0.54,0.51,0.47,0.46,0.45,0.41,0.40,0.38,0.37,0.36,0.33,0.36,0.64
9,,,,,,,,,,1.00,0.85,0.75,0.70,0.64,0.59,0.55,0.52,0.48,0.47,0.45,0.42,0.41,0.39,0.39,0.37,0.35,0.37,0.68
10,,,,,,,,,,,1.00,0.85,0.76,0.69,0.64,0.60,0.57,0.52,0.51,0.49,0.46,0.45,0.43,0.42,0.39,0.37,0.40,0.70
11,,,,,,,,,,,,1.00,0.84,0.75,0.67,0.63,0.58,0.53,0.51,0.51,0.48,0.47,0.45,0.43,0.41,0.39,0.40,0.68
12,,,,,,,,,,,,,1.00,0.85,0.76,0.71,0.66,0.58,0.56,0.55,0.52,0.50,0.48,0.48,0.45,0.44,0.44,0.72
13,,,,,,,,,,,,,,1.00,0.86,0.78,0.72,0.63,0.60,0.60,0.58,0.56,0.54,0.52,0.49,0.47,0.48,0.73
14,,,,,,,,,,,,,,,1.00,0.86,0.77,0.66,0.63,0.63,0.60,0.59,0.56,0.55,0.51,0.50,0.50,0.71
15,,,,,,,,,,,,,,,,1.00,0.85,0.74,0.68,0.67,0.64,0.62,0.59,0.59,0.55,0.55,0.56,0.75
16,,,,,,,,,,,,,,,,,1.00,0.80,0.73,0.72,0.69,0.68,0.64,0.65,0.62,0.59,0.58,0.70
17,,,,,,,,,,,,,,,,,,1.00,0.81,0.73,0.67,0.64,0.60,0.62,0.58,0.57,0.55,0.57
18,,,,,,,,,,,,,,,,,,,1.00,0.87,0.79,0.74,0.68,0.66,0.62,0.63,0.67,0.71
19,,,,,,,,,,,,,,,,,,,,1.00,0.88,0.81,0.74,0.70,0.66,0.67,0.71,0.75
20,,,,,,,,,,,,,,,,,,,,,1.00,0.90,0.83,0.78,0.72,0.72,0.72,0.76
21,,,,,,,,,,,,,,,,,,,,,,1.00,0.90,0.85,0.79,0.75,0.72,0.81
22,,,,,,,,,,,,,,,,,,,,,,,1.00,0.93,0.86,0.81,0.72,0.66
23,,,,,,,,,,,,,,,,,,,,,,,,1.00,0.90,0.81,0.71,0.55
24,,,,,,,,,,,,,,,,,,,,,,,,,1.00,0.83,0.74,0.49
25,,,,,,,,,,,,,,,,,,,,,,,,,,1.00,0.90,0.92
26,,,,,,,,,,,,,,,,,,,,,,,,,,,1.00,0.92
27,,,,,,,,,,,,,,,,,,,,,,,,,,,,1.00
\end{filecontents*}
\pgfplotstabletypeset[
          col sep = comma,
          every head row/.style={before row=\toprule,after row=\midrule},
          every last row/.style={after row=\bottomrule},
          display columns/0/.style={
            string type,
            column name={Layer},
          },
        ]{tables/qwen_diff_matrix.csv}
      \end{adjustbox}
    \end{subtable}\vfill\vspace{5pt}
    \begin{subtable}[t]{\textwidth}
      \centering
      \caption*{Diff-Vector CosSim}
      \begin{adjustbox}{max totalsize={\linewidth}{\textheight},center}
        \begin{filecontents*}{tables/qwen_token_matrix.csv}
i,0,1,2,3,4,5,6,7,8,9,10,11,12,13,14,15,16,17,18,19,20,21,22,23,24,25,26,27
0,1.00,0.31,0.33,0.29,0.24,0.23,0.18,0.14,0.12,0.11,0.11,0.10,0.10,0.09,0.08,0.07,0.07,0.05,0.04,0.04,0.03,0.03,0.04,0.02,0.02,0.02,0.04,0.12
1,,1.00,0.33,0.30,0.17,0.16,0.12,0.09,0.09,0.08,0.08,0.08,0.07,0.07,0.06,0.06,0.06,0.04,0.03,0.03,0.02,0.03,0.04,0.03,0.03,0.02,0.03,0.08
2,,,1.00,0.55,0.47,0.38,0.30,0.23,0.19,0.18,0.18,0.15,0.14,0.13,0.11,0.10,0.09,0.07,0.07,0.05,0.04,0.04,0.05,0.04,0.03,0.02,0.03,0.11
3,,,,1.00,0.56,0.44,0.32,0.26,0.23,0.22,0.21,0.18,0.17,0.16,0.14,0.13,0.11,0.08,0.08,0.07,0.05,0.05,0.06,0.05,0.05,0.03,0.05,0.12
4,,,,,1.00,0.51,0.36,0.29,0.26,0.24,0.23,0.19,0.18,0.16,0.14,0.13,0.12,0.09,0.08,0.07,0.06,0.06,0.06,0.05,0.05,0.03,0.05,0.13
5,,,,,,1.00,0.43,0.32,0.26,0.25,0.23,0.20,0.17,0.16,0.13,0.13,0.12,0.08,0.08,0.06,0.05,0.05,0.06,0.05,0.05,0.03,0.05,0.13
6,,,,,,,1.00,0.43,0.32,0.28,0.25,0.22,0.19,0.18,0.14,0.13,0.12,0.09,0.08,0.07,0.06,0.05,0.06,0.05,0.04,0.03,0.04,0.12
7,,,,,,,,1.00,0.48,0.40,0.34,0.28,0.24,0.21,0.17,0.16,0.14,0.10,0.09,0.08,0.06,0.06,0.07,0.06,0.05,0.03,0.05,0.14
8,,,,,,,,,1.00,0.47,0.37,0.30,0.25,0.22,0.18,0.16,0.14,0.10,0.09,0.07,0.05,0.06,0.06,0.06,0.05,0.03,0.05,0.14
9,,,,,,,,,,1.00,0.47,0.36,0.29,0.26,0.20,0.18,0.16,0.11,0.10,0.08,0.07,0.06,0.07,0.06,0.05,0.04,0.05,0.14
10,,,,,,,,,,,1.00,0.51,0.38,0.32,0.24,0.21,0.18,0.13,0.12,0.09,0.07,0.07,0.07,0.07,0.07,0.04,0.06,0.15
11,,,,,,,,,,,,1.00,0.49,0.39,0.29,0.26,0.21,0.13,0.12,0.10,0.07,0.07,0.07,0.08,0.07,0.04,0.07,0.14
12,,,,,,,,,,,,,1.00,0.56,0.40,0.31,0.24,0.16,0.13,0.10,0.07,0.07,0.08,0.08,0.07,0.04,0.07,0.15
13,,,,,,,,,,,,,,1.00,0.55,0.41,0.30,0.20,0.16,0.12,0.09,0.09,0.08,0.09,0.08,0.05,0.08,0.15
14,,,,,,,,,,,,,,,1.00,0.53,0.37,0.23,0.18,0.13,0.11,0.10,0.09,0.09,0.09,0.05,0.09,0.15
15,,,,,,,,,,,,,,,,1.00,0.55,0.35,0.27,0.19,0.15,0.12,0.12,0.13,0.12,0.06,0.12,0.16
16,,,,,,,,,,,,,,,,,1.00,0.44,0.29,0.19,0.13,0.11,0.10,0.11,0.10,0.05,0.10,0.15
17,,,,,,,,,,,,,,,,,,1.00,0.33,0.18,0.13,0.11,0.09,0.10,0.10,0.04,0.08,0.11
18,,,,,,,,,,,,,,,,,,,1.00,0.36,0.28,0.20,0.19,0.21,0.19,0.10,0.20,0.13
19,,,,,,,,,,,,,,,,,,,,1.00,0.42,0.27,0.24,0.26,0.23,0.13,0.24,0.14
20,,,,,,,,,,,,,,,,,,,,,1.00,0.41,0.36,0.35,0.30,0.18,0.26,0.15
21,,,,,,,,,,,,,,,,,,,,,,1.00,0.38,0.33,0.28,0.24,0.29,0.17
22,,,,,,,,,,,,,,,,,,,,,,,1.00,0.48,0.37,0.32,0.22,0.15
23,,,,,,,,,,,,,,,,,,,,,,,,1.00,0.68,0.17,0.22,0.10
24,,,,,,,,,,,,,,,,,,,,,,,,,1.00,0.17,0.22,0.07
25,,,,,,,,,,,,,,,,,,,,,,,,,,1.00,0.48,0.23
26,,,,,,,,,,,,,,,,,,,,,,,,,,,1.00,0.24
27,,,,,,,,,,,,,,,,,,,,,,,,,,,,1.00
\end{filecontents*}
\pgfplotstabletypeset[
          col sep = comma,
          every head row/.style={before row=\toprule,after row=\midrule},
          every last row/.style={after row=\bottomrule},
          display columns/0/.style={
            string type,
            column name={Layer},
          },
        ]{tables/qwen_token_matrix.csv}
      \end{adjustbox}
    \end{subtable}

  \endgroup
\end{table}

\begin{table}[H]
  \centering
  \caption{\textbf{Llama3.1-8B-It.} Raw scores for the plots in \Cref{fig:persistence_both} (bottom).}
  \label{tab:llama_persistence_raw}
  \begingroup
    \setlength{\tabcolsep}{4pt}
    \renewcommand{\arraystretch}{1.0}

    \begin{subtable}[t]{\textwidth}
      \centering
      \caption*{Diff-Diff CosSim}
      \begin{adjustbox}{max totalsize={\linewidth}{\textheight},center}
        \begin{filecontents*}{tables/llama_diff_matrix.csv}
i,0,1,2,3,4,5,6,7,8,9,10,11,12,13,14,15,16,17,18,19,20,21,22,23,24,25,26,27,28,29,30,31
0,1.00,0.84,0.68,0.55,0.46,0.38,0.34,0.31,0.30,0.29,0.26,0.24,0.27,0.34,0.35,0.37,0.40,0.42,0.43,0.46,0.44,0.46,0.46,0.46,0.45,0.45,0.46,0.46,0.44,0.42,0.53,0.57
1,,1.00,0.77,0.61,0.51,0.42,0.36,0.32,0.29,0.28,0.25,0.24,0.25,0.33,0.34,0.36,0.40,0.43,0.45,0.47,0.45,0.48,0.48,0.49,0.48,0.47,0.48,0.48,0.47,0.44,0.62,0.57
2,,,1.00,0.74,0.61,0.50,0.44,0.38,0.34,0.32,0.29,0.27,0.28,0.33,0.33,0.36,0.38,0.41,0.42,0.44,0.43,0.45,0.45,0.45,0.44,0.44,0.44,0.46,0.45,0.43,0.58,0.56
3,,,,1.00,0.78,0.62,0.51,0.44,0.38,0.35,0.31,0.29,0.31,0.37,0.39,0.40,0.45,0.46,0.47,0.49,0.48,0.50,0.50,0.51,0.50,0.49,0.50,0.50,0.50,0.47,0.54,0.58
4,,,,,1.00,0.75,0.61,0.50,0.44,0.38,0.33,0.30,0.31,0.43,0.44,0.47,0.51,0.52,0.53,0.55,0.53,0.56,0.56,0.58,0.56,0.55,0.56,0.55,0.54,0.50,0.53,0.53
5,,,,,,1.00,0.77,0.62,0.52,0.47,0.42,0.38,0.36,0.39,0.37,0.38,0.42,0.44,0.45,0.46,0.45,0.47,0.47,0.48,0.47,0.46,0.47,0.47,0.46,0.43,0.55,0.57
6,,,,,,,1.00,0.80,0.66,0.58,0.52,0.47,0.46,0.47,0.45,0.46,0.48,0.51,0.51,0.52,0.51,0.53,0.54,0.55,0.54,0.53,0.54,0.59,0.57,0.55,0.58,0.64
7,,,,,,,,1.00,0.77,0.66,0.55,0.48,0.45,0.51,0.50,0.52,0.54,0.56,0.57,0.58,0.56,0.59,0.59,0.60,0.59,0.57,0.60,0.59,0.58,0.54,0.53,0.56
8,,,,,,,,,1.00,0.82,0.68,0.58,0.53,0.51,0.48,0.47,0.51,0.53,0.53,0.54,0.52,0.54,0.54,0.54,0.53,0.51,0.52,0.52,0.51,0.48,0.48,0.55
9,,,,,,,,,,1.00,0.82,0.70,0.62,0.58,0.55,0.53,0.55,0.57,0.56,0.56,0.55,0.57,0.57,0.58,0.56,0.55,0.55,0.55,0.53,0.50,0.50,0.59
10,,,,,,,,,,,1.00,0.82,0.71,0.64,0.59,0.56,0.58,0.59,0.58,0.58,0.56,0.59,0.59,0.60,0.59,0.58,0.59,0.58,0.57,0.54,0.54,0.62
11,,,,,,,,,,,,1.00,0.85,0.76,0.67,0.60,0.60,0.59,0.57,0.55,0.53,0.56,0.56,0.56,0.54,0.53,0.53,0.52,0.51,0.49,0.50,0.59
12,,,,,,,,,,,,,1.00,0.83,0.71,0.64,0.64,0.63,0.61,0.60,0.58,0.60,0.60,0.60,0.58,0.56,0.57,0.56,0.55,0.52,0.52,0.58
13,,,,,,,,,,,,,,1.00,0.83,0.74,0.70,0.67,0.64,0.63,0.60,0.63,0.63,0.64,0.62,0.61,0.62,0.61,0.61,0.57,0.55,0.56
14,,,,,,,,,,,,,,,1.00,0.81,0.74,0.71,0.68,0.66,0.63,0.65,0.64,0.64,0.62,0.61,0.60,0.59,0.58,0.54,0.54,0.59
15,,,,,,,,,,,,,,,,1.00,0.88,0.80,0.75,0.70,0.66,0.65,0.63,0.62,0.59,0.57,0.57,0.56,0.55,0.53,0.55,0.59
16,,,,,,,,,,,,,,,,,1.00,0.86,0.79,0.74,0.69,0.67,0.65,0.62,0.59,0.57,0.57,0.56,0.55,0.52,0.59,0.59
17,,,,,,,,,,,,,,,,,,1.00,0.90,0.84,0.78,0.74,0.72,0.69,0.66,0.63,0.61,0.60,0.61,0.58,0.66,0.65
18,,,,,,,,,,,,,,,,,,,1.00,0.91,0.85,0.80,0.77,0.73,0.70,0.68,0.66,0.64,0.64,0.61,0.66,0.66
19,,,,,,,,,,,,,,,,,,,,1.00,0.91,0.85,0.80,0.77,0.73,0.70,0.67,0.66,0.65,0.63,0.66,0.62
20,,,,,,,,,,,,,,,,,,,,,1.00,0.91,0.86,0.82,0.77,0.74,0.71,0.69,0.68,0.65,0.71,0.69
21,,,,,,,,,,,,,,,,,,,,,,1.00,0.93,0.88,0.84,0.81,0.78,0.76,0.75,0.71,0.75,0.67
22,,,,,,,,,,,,,,,,,,,,,,,1.00,0.95,0.90,0.87,0.84,0.81,0.80,0.75,0.76,0.67
23,,,,,,,,,,,,,,,,,,,,,,,,1.00,0.94,0.89,0.86,0.83,0.82,0.76,0.77,0.70
24,,,,,,,,,,,,,,,,,,,,,,,,,1.00,0.92,0.88,0.85,0.83,0.77,0.78,0.68
25,,,,,,,,,,,,,,,,,,,,,,,,,,1.00,0.93,0.88,0.81,0.75,0.73,0.60
26,,,,,,,,,,,,,,,,,,,,,,,,,,,1.00,0.92,0.80,0.73,0.71,0.62
27,,,,,,,,,,,,,,,,,,,,,,,,,,,,1.00,0.90,0.81,0.74,0.70
28,,,,,,,,,,,,,,,,,,,,,,,,,,,,,1.00,0.86,0.81,0.66
29,,,,,,,,,,,,,,,,,,,,,,,,,,,,,,1.00,0.87,0.72
30,,,,,,,,,,,,,,,,,,,,,,,,,,,,,,,1.00,0.78
31,,,,,,,,,,,,,,,,,,,,,,,,,,,,,,,,1.00
\end{filecontents*}
\pgfplotstabletypeset[
          col sep = comma,
          every head row/.style={before row=\toprule,after row=\midrule},
          every last row/.style={after row=\bottomrule},
          display columns/0/.style={
            string type,
            column name={Layer},
          },
        ]{tables/llama_diff_matrix.csv}
      \end{adjustbox}
    \end{subtable}\vfill\vspace{5pt}
    \begin{subtable}[t]{\textwidth}
      \centering
      \caption*{Diff-Vector CosSim}
      \begin{adjustbox}{max totalsize={\linewidth}{\textheight},center}
        \begin{filecontents*}{tables/llama_token_matrix.csv}
i,0,1,2,3,4,5,6,7,8,9,10,11,12,13,14,15,16,17,18,19,20,21,22,23,24,25,26,27,28,29,30,31
0,1.00,0.17,0.13,0.08,0.05,0.03,0.03,0.03,0.03,0.01,0.01,0.01,0.03,0.05,0.05,0.05,0.05,0.05,0.04,0.04,0.04,0.03,0.03,0.04,0.03,0.03,0.02,0.01,-0.00,0.00,0.10,0.02
1,,1.00,0.18,0.10,0.06,0.04,0.03,0.03,0.02,0.01,0.01,0.01,0.03,0.05,0.05,0.05,0.05,0.06,0.04,0.04,0.04,0.03,0.03,0.04,0.03,0.03,0.02,0.01,0.00,0.00,0.15,0.01
2,,,1.00,0.20,0.11,0.07,0.06,0.05,0.04,0.02,0.02,0.02,0.03,0.05,0.05,0.05,0.05,0.06,0.05,0.04,0.05,0.04,0.04,0.05,0.04,0.03,0.03,0.02,0.00,0.00,0.13,0.01
3,,,,1.00,0.19,0.10,0.07,0.06,0.05,0.03,0.02,0.02,0.04,0.05,0.05,0.06,0.06,0.05,0.05,0.04,0.05,0.03,0.03,0.04,0.04,0.03,0.02,0.01,0.00,0.00,0.10,0.02
4,,,,,1.00,0.21,0.13,0.09,0.07,0.05,0.03,0.03,0.05,0.07,0.06,0.07,0.07,0.06,0.05,0.04,0.05,0.03,0.03,0.04,0.03,0.03,0.02,0.01,-0.01,0.00,0.06,0.01
5,,,,,,1.00,0.21,0.15,0.09,0.07,0.06,0.04,0.05,0.07,0.06,0.07,0.07,0.07,0.06,0.05,0.05,0.04,0.04,0.05,0.04,0.04,0.02,0.01,0.00,0.00,0.11,0.02
6,,,,,,,1.00,0.25,0.12,0.09,0.08,0.04,0.06,0.07,0.07,0.08,0.08,0.08,0.07,0.06,0.07,0.05,0.05,0.06,0.05,0.05,0.03,0.03,0.02,0.00,0.06,0.02
7,,,,,,,,1.00,0.24,0.17,0.12,0.08,0.09,0.10,0.08,0.09,0.09,0.08,0.07,0.05,0.06,0.04,0.04,0.05,0.04,0.04,0.02,0.02,0.00,0.01,0.05,0.02
8,,,,,,,,,1.00,0.28,0.16,0.10,0.12,0.10,0.09,0.09,0.10,0.09,0.08,0.06,0.07,0.05,0.05,0.06,0.05,0.04,0.02,0.02,0.01,0.01,0.06,0.02
9,,,,,,,,,,1.00,0.27,0.19,0.17,0.12,0.11,0.11,0.11,0.09,0.08,0.07,0.06,0.05,0.05,0.06,0.05,0.04,0.02,0.02,0.01,0.01,0.06,0.02
10,,,,,,,,,,,1.00,0.21,0.17,0.12,0.12,0.12,0.11,0.10,0.09,0.07,0.07,0.05,0.05,0.06,0.05,0.05,0.03,0.02,0.00,0.01,0.06,0.02
11,,,,,,,,,,,,1.00,0.24,0.20,0.16,0.14,0.12,0.10,0.09,0.08,0.07,0.06,0.06,0.06,0.05,0.04,0.02,0.03,0.01,0.02,0.07,0.02
12,,,,,,,,,,,,,1.00,0.25,0.21,0.17,0.15,0.12,0.10,0.09,0.09,0.07,0.06,0.07,0.06,0.05,0.03,0.03,0.01,0.02,0.07,0.02
13,,,,,,,,,,,,,,1.00,0.27,0.19,0.18,0.14,0.12,0.10,0.09,0.07,0.07,0.07,0.06,0.05,0.03,0.02,0.01,0.02,0.07,0.02
14,,,,,,,,,,,,,,,1.00,0.29,0.22,0.15,0.13,0.12,0.12,0.09,0.08,0.08,0.07,0.06,0.04,0.03,0.02,0.02,0.08,0.02
15,,,,,,,,,,,,,,,,1.00,0.28,0.21,0.17,0.15,0.14,0.12,0.10,0.10,0.09,0.07,0.05,0.04,0.03,0.03,0.10,0.02
16,,,,,,,,,,,,,,,,,1.00,0.29,0.25,0.19,0.17,0.15,0.13,0.13,0.11,0.08,0.06,0.05,0.04,0.04,0.15,0.03
17,,,,,,,,,,,,,,,,,,1.00,0.35,0.28,0.25,0.22,0.19,0.20,0.16,0.13,0.11,0.09,0.08,0.07,0.18,0.03
18,,,,,,,,,,,,,,,,,,,1.00,0.36,0.31,0.28,0.22,0.23,0.21,0.14,0.10,0.10,0.09,0.07,0.18,0.04
19,,,,,,,,,,,,,,,,,,,,1.00,0.36,0.31,0.29,0.23,0.21,0.16,0.11,0.10,0.10,0.08,0.18,0.05
20,,,,,,,,,,,,,,,,,,,,,1.00,0.44,0.33,0.32,0.31,0.19,0.15,0.12,0.13,0.08,0.20,0.05
21,,,,,,,,,,,,,,,,,,,,,,1.00,0.46,0.43,0.41,0.24,0.19,0.17,0.17,0.09,0.20,0.05
22,,,,,,,,,,,,,,,,,,,,,,,1.00,0.49,0.47,0.30,0.23,0.22,0.20,0.12,0.24,0.07
23,,,,,,,,,,,,,,,,,,,,,,,,1.00,0.57,0.34,0.28,0.23,0.21,0.12,0.24,0.06
24,,,,,,,,,,,,,,,,,,,,,,,,,1.00,0.38,0.32,0.29,0.25,0.12,0.25,0.06
25,,,,,,,,,,,,,,,,,,,,,,,,,,1.00,0.30,0.21,0.17,0.11,0.17,0.04
26,,,,,,,,,,,,,,,,,,,,,,,,,,,1.00,0.30,0.18,0.10,0.17,0.05
27,,,,,,,,,,,,,,,,,,,,,,,,,,,,1.00,0.26,0.15,0.20,0.10
28,,,,,,,,,,,,,,,,,,,,,,,,,,,,,1.00,0.27,0.30,0.07
29,,,,,,,,,,,,,,,,,,,,,,,,,,,,,,1.00,0.30,0.06
30,,,,,,,,,,,,,,,,,,,,,,,,,,,,,,,1.00,0.22
31,,,,,,,,,,,,,,,,,,,,,,,,,,,,,,,,1.00
\end{filecontents*}
\pgfplotstabletypeset[
          col sep = comma,
          every head row/.style={before row=\toprule,after row=\midrule},
          every last row/.style={after row=\bottomrule},
          display columns/0/.style={
            string type,
            column name={Layer},
          },
        ]{tables/llama_token_matrix.csv}
      \end{adjustbox}
    \end{subtable}

  \endgroup
\end{table}

\section{Steering Vector Persistence. Mean shift alignment on the last layer. Llama3.1-8B-It}
\label{appendix:persistence_last_diff_llama}

\begin{figure}[H]
  \centering
  \includegraphics[width=0.65\linewidth]{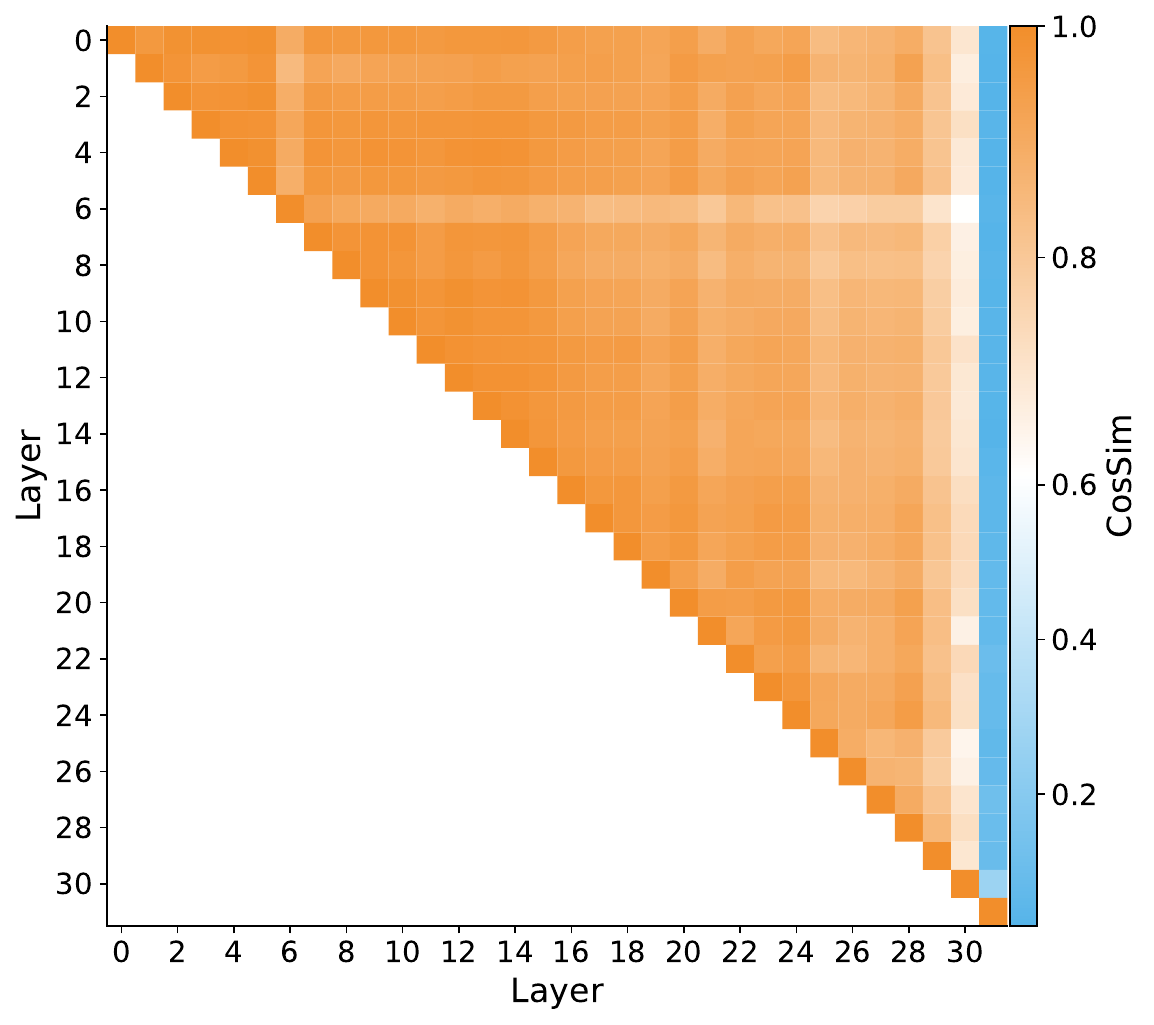}
\caption{
\textbf{Similarity of steering-induced unembedding biases (Llama3.1-8B-It).}
Each cell shows the cosine similarity between the average hidden-state shift at the final transformer layer induced by steering at layers $i$ and $j$. High similarity among $i,j<L$ indicates that steering from most layers produces a similar bias at the unembedding, largely independent of the injection point. In contrast, steering at the last layer directly yields a qualitatively different shift, implying a distinct mechanism.}
  \label{fig:persistence_last_diff_llama}
\end{figure}%
\section{Orthogonal Steering Vectors}
\label{appendix:qwen_ortho}

\begin{figure}[h]
  \centering
  \includegraphics[width=0.8\linewidth,page=1]{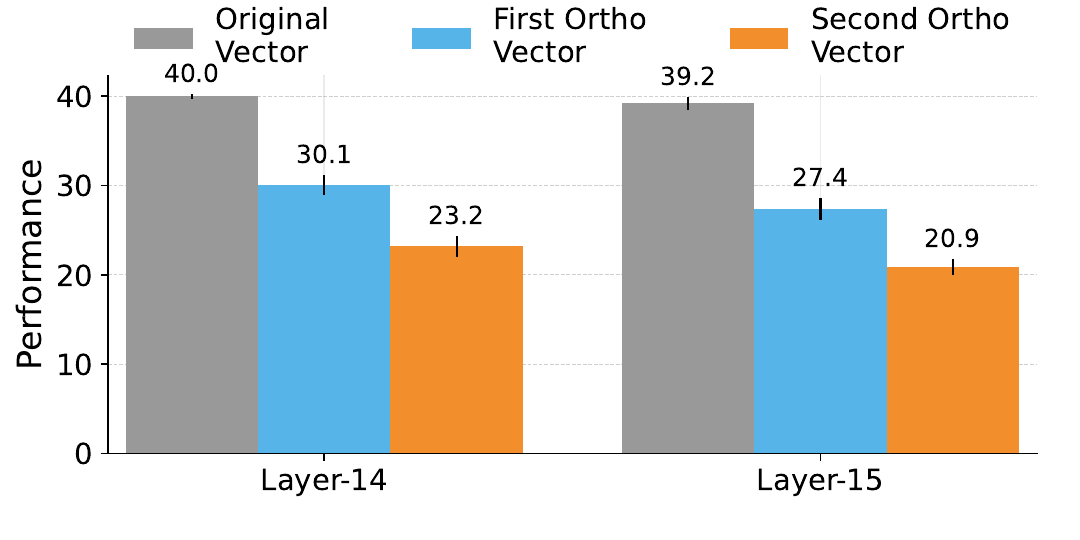}
  \caption{\textbf{Orthogonal steering vectors.} Performance of three mutually orthogonal steering vectors trained at layers 14 and 15. Contrary to the findings of \citet{ifound800} in our setup the orthogonal vectors do not achieve comparable performance: accuracy decreases monotonically as additional orthogonal vectors are trained.}

  \label{fig:qwen_ortho}
\end{figure}

\section{Last Layer. Logit Lens}
\label{appendix:last_layer_logit_lens}

\begin{table}[h]
  \centering
  \caption{\textbf{Last Layer -- logit-lens.} Cosine similarities and dot‐product scores between the last‐layer steering vector (trained in isolation) and the unembedding vectors of the top‐10 tokens for \texttt{Qwen2.5-Math-7B} and \texttt{Llama3.1-8B-It}.}
  \label{tab:last_layer_logit_lens}
  \small
  \setlength{\tabcolsep}{8pt}       
  \renewcommand{\arraystretch}{1.1} 

  \begin{subtable}[t]{\textwidth}
    \centering
    \caption*{Qwen2.5-Math-7B}
    \label{tab:last_layer_logit_lens_qwen}
    \begin{tabular}{lrrrrrrrrrr}
      \toprule
       & \texttt{To} & \texttt{]} & \texttt{To} & \texttt{So} & \texttt{\_to}
       & \texttt{\textbackslash} & \texttt{\}} & \texttt{For} & \texttt{.To} & \texttt{-to} \\
      \midrule
      Cos.\ Sim.\ & 0.37 & 0.16 & 0.16 & 0.15 & 0.14 & 0.14 & 0.13 & 0.13 & 0.12 & 0.12 \\
      Dot\ Prod.\ & 42.5 & 19.12 & 18.62 & 19.12 & 16.88 & 19.75 & 15.69 & 14.19 & 17.0 & 18.62 \\
      \bottomrule
    \end{tabular}
  \end{subtable}

  \vspace{1em}

  \begin{subtable}[t]{\textwidth}
    \centering
    \caption*{Llama3.1-8B-It}
    \label{tab:last_layer_logit_lens_llama}
    \begin{tabular}{lrrrrrrrrrr}
      \toprule
       & \texttt{final} & \texttt{Step} & \texttt{format} & \texttt{Final} & \texttt{final}
       & \texttt{final} & \texttt{Steps} & \texttt{Final} & \texttt{\_final} & \texttt{solution} \\
      \midrule
      Cos.\ Sim. & 0.12 & 0.11 & 0.09 & 0.09 & 0.08 & 0.08 & 0.08 & 0.08 & 0.08 & 0.08 \\
      Dot\ Prod. & 1.69 & 1.32 & 1.17 & 1.09 & 0.71 & 1.01 & 1.02 & 0.93 & 0.83 & 0.95 \\
      \bottomrule
    \end{tabular}
  \end{subtable}

\end{table}

\section{Last Layer Steering. Open-s1 dataset}
\label{appendix:token_to_opens1}

\begin{figure*}[h]
  \centering

  \begin{subfigure}[h]{\textwidth}
    \centering
    \begin{subfigure}[h]{0.48\textwidth}
      \centering
      \includegraphics[width=\linewidth,page=1]{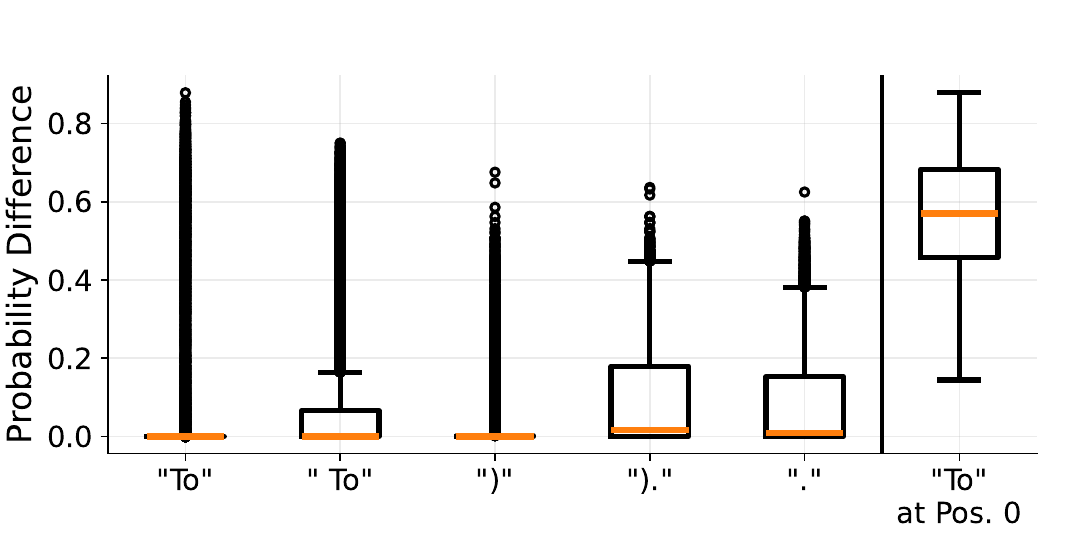}
      \caption*{}
    \end{subfigure}\hfill
    \begin{subfigure}[h]{0.48\textwidth}
      \centering
      \includegraphics[width=\linewidth,page=1]{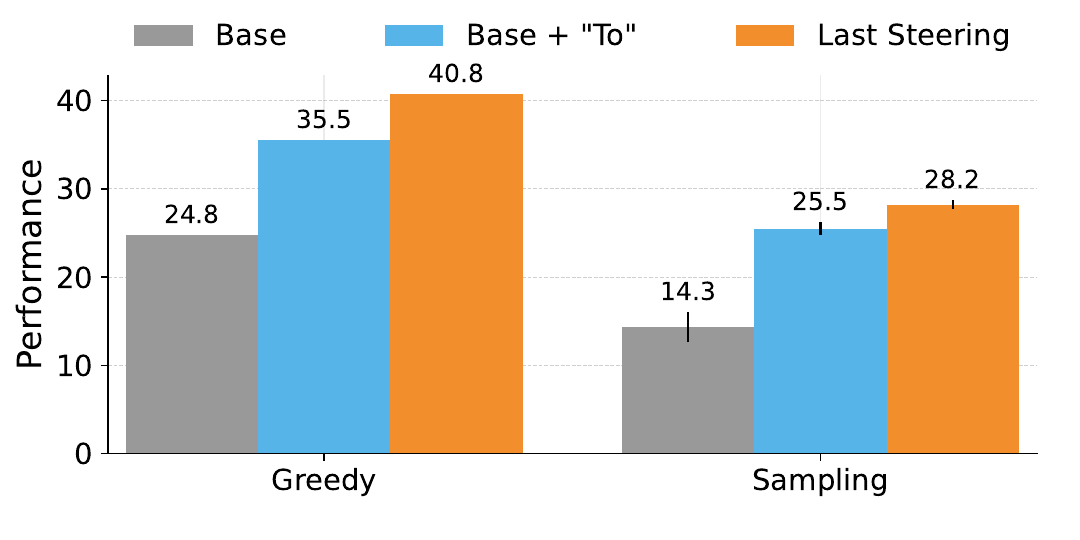}
      \caption*{}
    \end{subfigure}
    \vspace{-1.5em}
    \caption*{Qwen2.5-Math-7B}
  \end{subfigure}

  \vspace{0.9em}

  \begin{subfigure}[h]{\textwidth}
    \centering

    \begin{subfigure}[h]{0.48\textwidth}
      \centering
      \includegraphics[width=\linewidth,page=1]{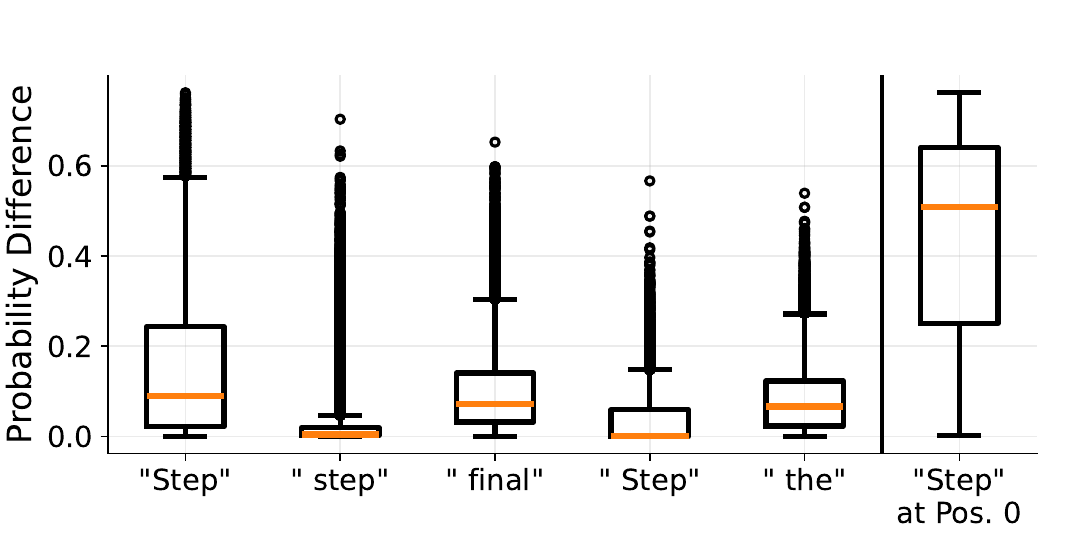}
      \caption*{}
    \end{subfigure}\hfill
    \begin{subfigure}[h]{0.48\textwidth}
      \centering
      \includegraphics[width=\linewidth,page=1]{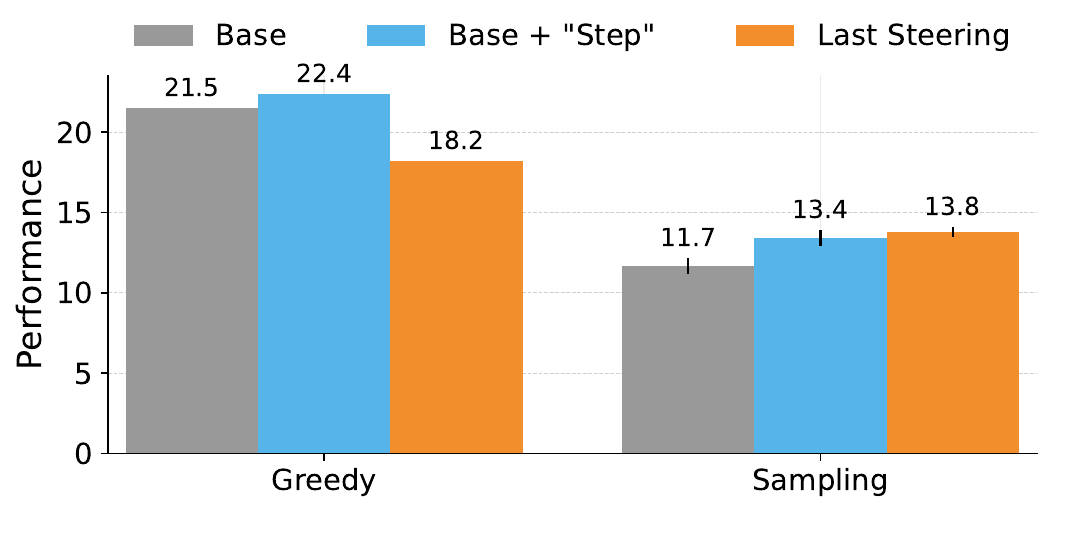}
      \caption*{}
    \end{subfigure}
    \vspace{-1.5em}
    \caption*{Llama3.1-8B-It}
  \end{subfigure}

  \caption{\textbf{Last-layer steering vector.}
\emph{Left}: distribution of token-level probability change induced by the last-layer vector over 1000 Open-s1 prompts. We include the top-5 tokens by maximum change and highlight the most affected token at the zeroth generation position.
\emph{Right}: prefixing this token to each prompt reproduces a substantial fraction of the vector's accuracy gain under both greedy decoding and sampling.}
  \label{fig:token_to_opens1}
\end{figure*}

\section{Value Steering Adds a Linear Term to MHA}
\label{appendix:v_1_proj}

The following derivation holds when we ignore the pre-attention LayerNorm (LN). While this is a strong assumption -- that LN does not alter the steering vector’s trajectory -- the experiment in \Cref{section:pre_last_layer} shows that a post-attention steering vector attains the same performance as a pre-attention one, indicating that the pre-attention vector indeed does not act through attention.

\textbf{Claim.} Let $U\in\mathbb R^{T\times d_{\text{model}}}$ and define the (row-wise) attention
\[
A(U)=\operatorname{Softmax}\!\left(\frac{UW_i^{Q}(UW_i^{K})^{\top}}{\sqrt{d_k}}\right)\in\mathbb R^{T\times T},
\quad A(U)\mathbf 1=\mathbf 1 .
\]
For head $i$,
\[
H_i(U)=A(U)\,U W_i^{V}.
\]
Let a steering vector $s\in\mathbb R^{d_{\text{model}}}$ be added to the \emph{values} of head $i$ for every token, and set
$S=\mathbf 1 s^{\top}\in\mathbb R^{T\times d_{\text{model}}}$. Then
\[
\begin{aligned}
H_i^{(+s)}(U)
&=A(U)\,(U+S)W_i^{V} \\
&=A(U)UW_i^{V}+A(U)SW_i^{V} \\
&=H_i(U)+SW_i^{V} \qquad (\text{since }A(U)\mathbf 1=\mathbf 1).
\end{aligned}
\]
Writing $W^{O}=\begin{bmatrix}W_1^{O}\\ \cdots \\ W_h^{O}\end{bmatrix}$ by heads, the multi-head output satisfies
\[
\boxed{\ \mathrm{MHA}(U+S)=\mathrm{MHA}(U)+S\,W_i^{V}W_i^{O}\ }
\]
and is independent of the attention pattern. 

\section{Unnormalized Transfer Performance}
\label{appendix:transfer_raw}

\begin{table}[htbp]
\centering
\caption{\textbf{Transferability of steering vectors within the Qwen2.5 family}. Each cell shows the mean performance change when the steering vector trained for the \emph{Donor} model is applied to the \emph{Recipient} model. The "None" column denotes the non-trained models' performance. Llama3.1-8B$^*$ denotes the transferring experiments where we used "Qwen-Math" template for Llama3.1-8B-Instruct as in the main experiments.}
\label{tab:transfer_raw}

\begin{tabular}{llcccc}
\toprule
\multicolumn{2}{c}{} & \multicolumn{4}{c}{\textbf{Donor}} \\
\cmidrule(lr){3-6}
\textbf{Family} & \textbf{Recipient} & \textbf{None} & \textbf{Base} & \textbf{Instruct} & \textbf{Math} \\
\midrule
\multirow{3}{*}{Qwen2.5-1.5B}
& Base       & 0.52 $\pm$ 0.12 & 22.72 $\pm$ 0.53 & 9.02 $\pm$ 1.66 & 7.57 $\pm$ 0.67 \\
& Instruct    & 13.44 $\pm$ 1.17 & 23.22 $\pm$ 0.35 & 23.82 $\pm$ 0.28 & 16.64 $\pm$ 0.27 \\
& Math  & 11.33 $\pm$ 1.09 & 19.48 $\pm$ 1.18 & 16.09 $\pm$ 1.48 & 34.11 $\pm$ 0.28 \\
\midrule
\multirow{3}{*}{Qwen2.5-7B}
& Base          & 12.04 $\pm$ 5.85 & 36.44 $\pm$ 0.15 & 20.78 $\pm$ 3.43 & 30.0 $\pm$ 0.14 \\
& Instruct       & 35.82 $\pm$ 0.14 & 37.51 $\pm$ 0.44 & 38.89 $\pm$ 0.27 & 34.78 $\pm$ 0.07 \\
& Math     & 14.33 $\pm$ 1.75 & 23.42 $\pm$ 1.76 & 15.84 $\pm$ 1.96 & 42.82 $\pm$ 0.25 \\
\midrule
\multirow{2}{*}{Llama3.1-8B}
& Base       & 0.91 $\pm$ 0.11 & 9.18 $\pm$ 0.22 & 3.24 $\pm$ 0.27 & --- \\
& Instruct    & 3.33 $\pm$ 0.16 & 17.03 $\pm$ 0.34 & 21.76 $\pm$ 0.29 & --- \\
\midrule
\multirow{2}{*}{Llama3.1-8B$^*$}
& Base       & 0.91 $\pm$ 0.11 & 9.18 $\pm$ 0.22 & 0.95 $\pm$ 0.11 & --- \\
& Instruct    & 11.81 $\pm$ 0.43 & 11.66 $\pm$ 0.46 & 26.14 $\pm$ 0.43 & --- \\
\bottomrule
\end{tabular}
\end{table}

\begin{table}[htbp]
\centering
\caption{\textbf{Transferability of steering vectors within the Qwen2.5 family -- AIME24}. Each cell shows the performance change when the steering vector trained for the \emph{Donor} model is applied to the \emph{Recipient} model. The "None" column denotes the non-trained models' performance. Llama3.1-8B$^*$ denotes the transferring experiments where we used "Qwen-Math" template for Llama3.1-8B-Instruct as in the main experiments.}
\label{tab:transfer_raw_aime24}

\begin{tabular}{llcccc}
\toprule
\multicolumn{2}{c}{} & \multicolumn{4}{c}{\textbf{Donor}} \\
\cmidrule(lr){3-6}
\textbf{Family} & \textbf{Recipient} & \textbf{None} & \textbf{Base} & \textbf{Instruct} & \textbf{Math} \\
\midrule
\multirow{3}{*}{Qwen2.5-1.5B}
& Base       & 0.0 $\pm$ 0.0 & 2.81 $\pm$ 0.09 & 1.70 $\pm$ 0.10 & 0.73 $\pm$ 0.0 \\
& Instruct    & 0.83 $\pm$ 0.09 & 2.60 $\pm$ 0.0 & 3.30 $\pm$ 0.13 & 0.87 $\pm$ 0.05 \\
& Math  & 2.92 $\pm$ 0.0 & 5.21 $\pm$ 0.09 & 4.83 $\pm$ 0.05 & 12.08 $\pm$ 0.0 \\
\midrule
\multirow{3}{*}{Qwen2.5-7B}
& Base          & 4.44 $\pm$ 6.29 & 12.78 $\pm$ 0.05 & 6.77 $\pm$ 0.0 & 8.78 $\pm$ 0.13 \\
& Instruct       & 11.11 $\pm$ 0.18 & 13.65 $\pm$ 0.0 & 14.79 $\pm$ 0.0 & 10.87 $\pm$ 0.13 \\
& Math     & 6.70 $\pm$ 0.18 & 11.11 $\pm$ 0.13 & 8.06 $\pm$ 0.18 & 24.31 $\pm$ 0.20 \\
\midrule
\multirow{2}{*}{Llama3.1-8B}
& Base       & 0.10 $\pm$ 0.0 & 0.28 $\pm$ 0.13 & 0.24 $\pm$ 0.05 & --- \\
& Instruct    & 0.45 $\pm$ 0.10 & 1.08 $\pm$ 0.05 & 6.63 $\pm$ 0.05 & --- \\
\bottomrule
\end{tabular}
\end{table}

\begin{table}[htbp]
\centering
\caption{\textbf{Transferability of steering vectors within the Qwen2.5 family -- AIME25}. Each cell shows the performance change when the steering vector trained for the \emph{Donor} model is applied to the \emph{Recipient} model. The "None" column denotes the non-trained models' performance. Llama3.1-8B$^*$ denotes the transferring experiments where we used "Qwen-Math" template for Llama3.1-8B-Instruct as in the main experiments.}
\label{tab:transfer_raw_aime25}

\begin{tabular}{llcccc}
\toprule
\multicolumn{2}{c}{} & \multicolumn{4}{c}{\textbf{Donor}} \\
\cmidrule(lr){3-6}
\textbf{Family} & \textbf{Recipient} & \textbf{None} & \textbf{Base} & \textbf{Instruct} & \textbf{Math} \\
\midrule
\multirow{3}{*}{Qwen2.5-1.5B}
& Base       & 0.10 $\pm$ 0.00 & 1.46 $\pm$ 0.00 & 1.35 $\pm$ 0.09 & 0.10 $\pm$ 0.00 \\
& Instruct   & 0.31 $\pm$ 0.09 & 1.60 $\pm$ 0.05 & 2.15 $\pm$ 0.05 & 0.83 $\pm$ 0.09 \\
& Math       & 1.81 $\pm$ 0.10 & 2.53 $\pm$ 0.05 & 2.57 $\pm$ 0.13 & 8.12 $\pm$ 0.09 \\
\midrule
\multirow{3}{*}{Qwen2.5-7B}
& Base       & 0.00 $\pm$ 0.00 & 7.12 $\pm$ 0.32 & 3.54 $\pm$ 0.15 & 5.49 $\pm$ 0.05 \\
& Instruct   & 6.74 $\pm$ 0.13 & 7.19 $\pm$ 0.09 & 8.72 $\pm$ 0.05 & 4.38 $\pm$ 0.09 \\
& Math       & 1.74 $\pm$ 0.13 & 4.41 $\pm$ 0.18 & 3.23 $\pm$ 0.00 & 12.64 $\pm$ 0.26 \\
\midrule
\multirow{2}{*}{Llama3.1-8B}
& Base       & 0.00 $\pm$ 0.00 & 0.28 $\pm$ 0.05 & 0.10 $\pm$ 0.00 & --- \\
& Instruct   & 0.07 $\pm$ 0.05 & 0.62 $\pm$ 0.00 & 1.88 $\pm$ 0.15 & --- \\
\bottomrule
\end{tabular}
\end{table}

\begin{table}[htbp]
\centering
\caption{\textbf{Transferability of steering vectors within the Qwen2.5 family -- AMC-23}. Each cell shows the performance change when the steering vector trained for the \emph{Donor} model is applied to the \emph{Recipient} model. The "None" column denotes the non-trained models' performance. Llama3.1-8B$^*$ denotes the transferring experiments where we used "Qwen-Math" template for Llama3.1-8B-Instruct as in the main experiments.}
\label{tab:transfer_raw_amc23}

\begin{tabular}{llcccc}
\toprule
\multicolumn{2}{c}{} & \multicolumn{4}{c}{\textbf{Donor}} \\
\cmidrule(lr){3-6}
\textbf{Family} & \textbf{Recipient} & \textbf{None} & \textbf{Base} & \textbf{Instruct} & \textbf{Math} \\
\midrule
\multirow{3}{*}{Qwen2.5-1.5B}
& Base       & 0.73 $\pm$ 0.10 & 33.91 $\pm$ 0.11 & 11.61 $\pm$ 0.13 & 8.49 $\pm$ 0.21 \\
& Instruct   & 16.64 $\pm$ 0.36 & 33.91 $\pm$ 0.17 & 31.82 $\pm$ 0.13 & 21.90 $\pm$ 0.10 \\
& Math       & 18.41 $\pm$ 0.16 & 32.08 $\pm$ 0.39 & 26.54 $\pm$ 0.21 & 48.96 $\pm$ 0.16 \\
\midrule
\multirow{3}{*}{Qwen2.5-7B}
& Base       & 17.50 $\pm$ 12.42 & 49.90 $\pm$ 0.13 & 32.45 $\pm$ 0.27 & 41.02 $\pm$ 0.33 \\
& Instruct   & 50.29 $\pm$ 0.16 & 50.39 $\pm$ 0.17 & 54.24 $\pm$ 0.15 & 46.25 $\pm$ 0.33 \\
& Math       & 21.30 $\pm$ 0.38 & 35.60 $\pm$ 0.42 & 25.57 $\pm$ 0.37 & 62.14 $\pm$ 0.30 \\
\midrule
\multirow{2}{*}{Llama3.1-8B}
& Base       & 0.94 $\pm$ 0.11 & 8.83 $\pm$ 0.06 & 2.81 $\pm$ 0.39 & --- \\
& Instruct   & 4.84 $\pm$ 0.34 & 22.60 $\pm$ 0.22 & 24.92 $\pm$ 0.19 & --- \\
\bottomrule
\end{tabular}
\end{table}

\begin{table}[htbp]
\centering
\caption{\textbf{Transferability of steering vectors within the Qwen2.5 family -- MATH-500}. Each cell shows the performance change when the steering vector trained for the \emph{Donor} model is applied to the \emph{Recipient} model. The "None" column denotes the non-trained models' performance. Llama3.1-8B$^*$ denotes the transferring experiments where we used "Qwen-Math" template for Llama3.1-8B-Instruct as in the main experiments.}
\label{tab:transfer_raw_math500}

\begin{tabular}{llcccc}
\toprule
\multicolumn{2}{c}{} & \multicolumn{4}{c}{\textbf{Donor}} \\
\cmidrule(lr){3-6}
\textbf{Family} & \textbf{Recipient} & \textbf{None} & \textbf{Base} & \textbf{Instruct} & \textbf{Math} \\
\midrule
\multirow{3}{*}{Qwen2.5-1.5B}
& Base       & 1.20 $\pm$ 0.28 & 56.80 $\pm$ 0.75 & 20.80 $\pm$ 5.58 & 21.87 $\pm$ 2.19 \\
& Instruct   & 39.93 $\pm$ 5.32 & 57.60 $\pm$ 1.18 & 61.13 $\pm$ 1.04 & 46.33 $\pm$ 1.24 \\
& Math       & 24.00 $\pm$ 4.14 & 44.40 $\pm$ 3.10 & 31.87 $\pm$ 5.03 & 71.47 $\pm$ 0.93 \\
\midrule
\multirow{3}{*}{Qwen2.5-7B}
& Base       & 30.00 $\pm$ 11.75 & 74.07 $\pm$ 0.62 & 48.00 $\pm$ 11.54 & 66.33 $\pm$ 2.35 \\
& Instruct   & 74.40 $\pm$ 1.18 & 76.20 $\pm$ 0.71 & 78.40 $\pm$ 1.56 & 74.80 $\pm$ 0.59 \\
& Math       & 37.67 $\pm$ 6.09 & 54.00 $\pm$ 4.14 & 37.80 $\pm$ 7.11 & 80.20 $\pm$ 0.33 \\
\midrule
\multirow{2}{*}{Llama3.1-8B}
& Base       & 1.87 $\pm$ 0.25 & 26.13 $\pm$ 1.06 & 11.33 $\pm$ 1.47 & --- \\
& Instruct   & 7.33 $\pm$ 0.66 & 44.47 $\pm$ 0.41 & 55.00 $\pm$ 0.33 & --- \\
\bottomrule
\end{tabular}
\end{table}

\begin{table}[htbp]
\centering
\caption{\textbf{Transferability of steering vectors within the Qwen2.5 family -- Minerva-Math}. Each cell shows the performance change when the steering vector trained for the \emph{Donor} model is applied to the \emph{Recipient} model. The "None" column denotes the non-trained models' performance. Llama3.1-8B$^*$ denotes the transferring experiments where we used "Qwen-Math" template for Llama3.1-8B-Instruct as in the main experiments.}
\label{tab:transfer_raw_minerva}

\begin{tabular}{llcccc}
\toprule
\multicolumn{2}{c}{} & \multicolumn{4}{c}{\textbf{Donor}} \\
\cmidrule(lr){3-6}
\textbf{Family} & \textbf{Recipient} & \textbf{None} & \textbf{Base} & \textbf{Instruct} & \textbf{Math} \\
\midrule
\multirow{3}{*}{Qwen2.5-1.5B}
& Base       & 0.61 $\pm$ 0.17 & 18.75 $\pm$ 2.38 & 7.60 $\pm$ 2.29 & 5.27 $\pm$ 0.35 \\
& Instruct   & 9.68 $\pm$ 1.05 & 19.49 $\pm$ 1.08 & 20.59 $\pm$ 1.83 & 13.24 $\pm$ 0.79 \\
& Math       & 5.39 $\pm$ 1.71 & 9.56 $\pm$ 2.96 & 9.44 $\pm$ 2.72 & 27.70 $\pm$ 1.05 \\
\midrule
\multirow{3}{*}{Qwen2.5-7B}
& Base       & 7.11 $\pm$ 2.88 & 37.13 $\pm$ 0.60 & 12.99 $\pm$ 2.44 & 25.74 $\pm$ 1.50 \\
& Instruct   & 35.54 $\pm$ 1.42 & 37.75 $\pm$ 1.54 & 35.54 $\pm$ 0.87 & 34.68 $\pm$ 0.46 \\
& Math       & 8.33 $\pm$ 1.51 & 11.52 $\pm$ 2.55 & 7.84 $\pm$ 0.96 & 34.44 $\pm$ 1.76 \\
\midrule
\multirow{2}{*}{Llama3.1-8B}
& Base       & 1.47 $\pm$ 0.52 & 12.62 $\pm$ 0.35 & 3.19 $\pm$ 0.69 & --- \\
& Instruct   & 4.41 $\pm$ 0.60 & 18.87 $\pm$ 1.05 & 21.20 $\pm$ 1.48 & --- \\
\bottomrule
\end{tabular}
\end{table}

\begin{table}[htbp]
\centering
\caption{\textbf{Transferability of steering vectors within the Qwen2.5 family -- OlympiadBench}. Each cell shows the performance change when the steering vector trained for the \emph{Donor} model is applied to the \emph{Recipient} model. The "None" column denotes the non-trained models' performance. Llama3.1-8B$^*$ denotes the transferring experiments where we used "Qwen-Math" template for Llama3.1-8B-Instruct as in the main experiments.}
\label{tab:transfer_raw_olympiad}

\begin{tabular}{llcccc}
\toprule
\multicolumn{2}{c}{} & \multicolumn{4}{c}{\textbf{Donor}} \\
\cmidrule(lr){3-6}
\textbf{Family} & \textbf{Recipient} & \textbf{None} & \textbf{Base} & \textbf{Instruct} & \textbf{Math} \\
\midrule
\multirow{3}{*}{Qwen2.5-1.5B}
& Base       & 0.44 $\pm$ 0.32 & 22.62 $\pm$ 0.55 & 11.06 $\pm$ 3.98 & 8.99 $\pm$ 1.48 \\
& Instruct   & 13.23 $\pm$ 1.14 & 24.10 $\pm$ 0.60 & 24.00 $\pm$ 1.05 & 16.64 $\pm$ 0.18 \\
& Math       & 15.46 $\pm$ 0.85 & 23.11 $\pm$ 1.19 & 21.28 $\pm$ 1.39 & 36.35 $\pm$ 0.69 \\
\midrule
\multirow{3}{*}{Qwen2.5-7B}
& Base       & 13.19 $\pm$ 6.40 & 37.63 $\pm$ 0.44 & 20.94 $\pm$ 6.50 & 32.64 $\pm$ 1.84 \\
& Instruct   & 36.84 $\pm$ 0.18 & 39.90 $\pm$ 0.74 & 41.68 $\pm$ 0.30 & 37.73 $\pm$ 0.55 \\
& Math       & 10.22 $\pm$ 2.94 & 23.85 $\pm$ 4.39 & 12.54 $\pm$ 3.34 & 43.21 $\pm$ 0.14 \\
\midrule
\multirow{2}{*}{Llama3.1-8B}
& Base       & 0.54 $\pm$ 0.14 & 7.21 $\pm$ 0.28 & 1.78 $\pm$ 0.64 & --- \\
& Instruct   & 2.86 $\pm$ 0.62 & 14.52 $\pm$ 0.74 & 20.94 $\pm$ 0.07 & --- \\
\bottomrule
\end{tabular}
\end{table}

\newpage
\section{Chat Templates}
\label{appendix:chat_templates}

Following \citet{liu2025understanding}, we used two chat templates. For models that support special chat-template tokens, we adopted the \emph{Qwen-Math} template; special tokens for \texttt{Qwen2.5-Math-7B} are shown as a representative example. For \texttt{Llama3.1-8B} -- which does not include pretrained special chat-template tokens -- we used the \emph{R1} template.

\begin{tcolorbox}[title=Chat Template -- Qwen-Math]
\ttfamily
<|im\_start|>system
Please reason step by step, and put your final answer within \textbackslash boxed\{\}.<|im\_end|>
<|im\_start|>user
TASK<|im\_end|>
<|im\_start|>assistant
\end{tcolorbox}

\begin{tcolorbox}[title=Chat Template -- R1]
\ttfamily
A conversation between User and Assistant. The User asks a question, and the Assistant solves it. The Assistant first thinks about the reasoning process in the mind and then provides the User with the answer. The reasoning process is enclosed within <think> </think> and answer is enclosed within <answer> </answer> tags, respectively, i.e., <think> reasoning process here </think> <answer> answer here </answer>.
\textbackslash nUser: TASK\textbackslash nAssistant: <think>
\end{tcolorbox}

\section{LLM Use}
\label{sec:llm_use}

We used ChatGPT to check grammar and clarity during the writing of this paper.


\end{document}